\pdfoutput=1

\documentclass{article}

\PassOptionsToPackage{numbers, compress}{natbib}

\usepackage[preprint]{neurips_2025}

\usepackage[utf8]{inputenc} 
\usepackage[T1]{fontenc}    
\usepackage{hyperref}       
\usepackage{url}            
\usepackage{booktabs}       
\usepackage{amsfonts}       
\usepackage{nicefrac}       
\usepackage{microtype}      
\usepackage{xcolor}         
\usepackage{xpatch}

\usepackage{multirow}
\usepackage{tcolorbox}
\usepackage{wrapfig}
\usepackage{booktabs}
\usepackage{colortbl}
\usepackage{amsmath}
\usepackage{float}

\usepackage{xpatch}
\makeatletter
\xapptocmd{\NAT@bibsetnum}{\setlength{\leftmargin}{0pt}\setlength{\itemindent}{\labelwidth}\addtolength{\itemindent}{\labelsep}}{}{}
\makeatother

\title{Multi-modal Retrieval Augmented Multi-modal Generation:
Datasets, Evaluation Metrics and Strong Baselines
}

\author{
  \textbf{Zi-Ao Ma\textsuperscript{1}},
  \textbf{Tian Lan\textsuperscript{1}},
  \textbf{Rong-Cheng Tu\textsuperscript{2}},
  \textbf{Yong Hu\textsuperscript{3}},
\\
  \textbf{Yu-Shi Zhu\textsuperscript{1}},
  \textbf{Tong Zhang\textsuperscript{1}},
  \textbf{Heyan Huang\textsuperscript{1}},
  \textbf{Zhijing Wu\textsuperscript{1}}\thanks{\quad Corresponding author},
  \textbf{Xian-Ling Mao\textsuperscript{1}}
\\
  \textsuperscript{1}School of Computer Science and Technology, Beijing Institute of Technology, China
\\
  \textsuperscript{2}Nanyang Technological University, Singapore,
  \textsuperscript{3}WeChat AI, Tencent Inc., China
\\
    \texttt{\{maziaoylwt,lantiangmftby\}@gmail.com,} \texttt{rongcheng.tu@ntu.edu.sg}
\\
    \texttt{rightyonghu@tencent.com,}
    \texttt{wuzhijing.joyce@gmail.com,}
    \texttt{maoxl@bit.edu.cn}
\\
\url{https://github.com/maziao/M2RAG}
}

\begin{document}

\maketitle

\begin{abstract}
We present a systematic investigation of Multi-modal Retrieval Augmented Multi-modal Generation (M$^2$RAG), a novel task that enables foundation models to process multi-modal web content and generate multi-modal responses, which exhibits better information density and readability.
Despite its potential impact, M$^2$RAG remains understudied, lacking comprehensive analysis and high-quality data resources.
To address this gap, we establish a comprehensive benchmark through a rigorous data curation pipeline, and employ text-modal metrics and multi-modal metrics based on foundation models for evaluation.
We further propose several strategies for foundation models to process M$^2$RAG task effectively and construct a training set by filtering high-quality samples using our designed metrics.
Our extensive experiments demonstrate the reliability of our proposed metrics, a landscape of model performance within our designed strategies, and show that our fine-tuned 7B-8B models outperform the GPT-4o model and approach the state-of-the-art OpenAI o3-mini. 
Additionally, we perform fine-grained analyses across diverse domains and validate the effectiveness of our designs in data curation pipeline. All resources, including codes, datasets, and model weights, will be publicly released.
\end{abstract}

\section{Introduction}

\begin{figure}[h]
    \centering
    \includegraphics[width=1.0\linewidth]{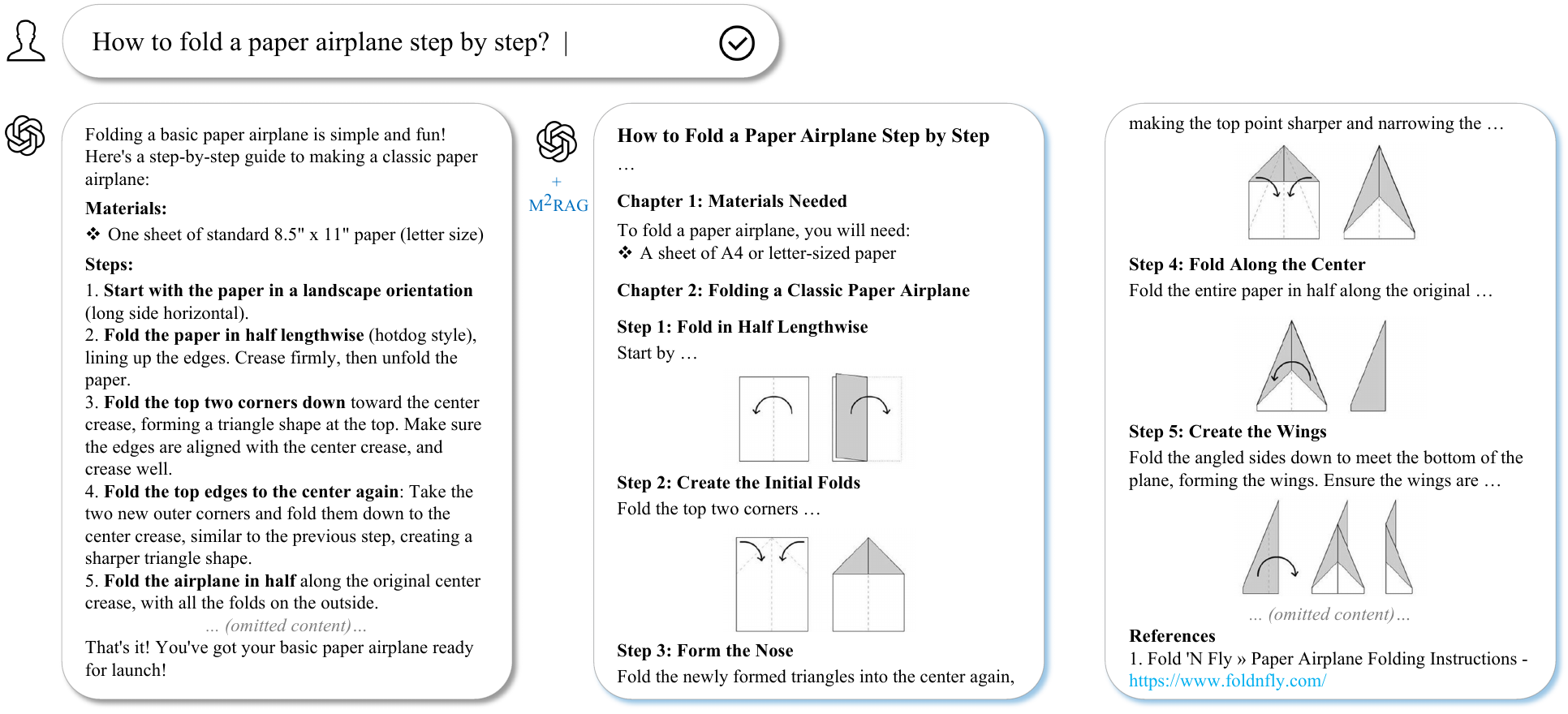}
    \caption{A typical comparison between naive RAG (upper) and our proposed M$^2$RAG (lower). The generative model is GPT-4o in this case.}
    \label{fig:typical_case}
    \vspace{-10pt}
\end{figure}

Retrieval Augmented Generation (RAG)~\cite{lan2023copyneed,sun2024block,li2022survey} and its multi-modal extensions ~\cite{chen2022murag,joshi2024robust,ye2024mplug} enhance foundation models by incorporating external knowledge and multi-modal data. While these methods improve response quality, they remain limited to textual outputs, which may fail to provide sufficient clarity in some scenarios, such as instructional guides or spatial reasoning. For instance, as shown in Figure~\ref{fig:typical_case}, a text-only response explaining paper airplane folding steps may be difficult for users to follow without accompanying visual illustrations.

As the saying goes, \textit{A single image is worth a thousand words.} Visual elements can significantly enhance comprehension, especially in instructional and knowledge-intensive domains where textual descriptions alone may be insufficient. Motivated by this, the Multi-modal Retrieval-Augmented Multi-modal Generation (M$^2$RAG) has been introduced ~\cite{zhu2024murar,wu2023next,li2024textbind}, a novel task that retrieves and integrates multi-modal content to produce responses with a mixed layout of text and image. By integrating relevant images into textual responses, M$^2$RAG enhances comprehension and usability across various application scenarios. As illustrated in Figure~\ref{fig:typical_case}, M$^2$RAG not only generates textual explanations for step-by-step folding a paper airplane but also seamlessly incorporates illustrative images where necessary, significantly improving user understanding.
Compared to existing RAG methods, M$^2$RAG introduces new challenges in multi-modal reasoning and synthesis. It requires models to achieve a deeper understanding of multi-modal data, accurately capture content and cross-modal relationships, and generate coherent responses that seamlessly integrate text and images~\cite{zhu2024murar}.

However, given the early research stage, there is a lack of systematic analysis and high-quality data for M$^2$RAG task. 
To fill this gap, we make the following contributions: 
(1) \textbf{Benchmark:} A comprehensive benchmark with 10 topics constructed by our rigorous data curation pipeline;
(2) \textbf{Evaluation Metric:} A suite of reliable text-only and multi-modal evaluation metrics based on Large Language Models (LLMs) and Multi-modal LLMs (MLLMs);
(3) \textbf{Generation Strategy:} Two generation strategies for effectively tackling M$^2$RAG task: single-stage and multi-stage approaches;
and (4) \textbf{Training Dataset:} A training dataset constructed by filtering high-quality samples using our designed multi-modal evaluation metrics, which is used to improve the performance of 7B-8B LLMs and MLLMs.

Based on our extensive experiments and automatic evaluation of advanced LLMs and MLLMs, we present four key findings:
(1) \textbf{Metric Reliability:} The designed evaluation metrics exhibit strong correlations with human judgments, even outperform inner-correlation of human annotators in evaluating image helpfulness dimension. This validates the reliability of our automatic evaluation framework;
(2) \textbf{Model Performance Analysis:} Experimental results on 12 advanced LLMs and MLLMs provides valuable insights for M$^2$RAG task. For example, LLMs consistently outperforms MLLMs, revealing MLLM's limited multi-modal understanding and generation abilities. Besides, multi-stage strategy integrates more relevant images into responses, exhibiting better overall quality;
(3) \textbf{Effectiveness of Training Dataset:} Fine-tuning 7B-8B LLMs and MLLMs on our curated dataset yields performance that exceeds GPT-4o, underscoring the quality and effectiveness of our training dataset;
and (4) \textbf{Data Curation Benefits:} Ablation studies quantify cross-domain performance variations and verify the positive contributions of our designs in benchmark data curation pipeline, including detailed context for images and the inclusion of the auxiliary images.
These observations and phenomena promote an in-depth understanding of M$^2$RAG. We hope our data resources and these discoveries could spur future research in this field.

\section{Related Work}

\paragraph{Text-modal RAG}
RAG~\cite{li2022survey} has been widely used to improve the generation quality of language models by incorporating external knowledge~\cite{lan2023copyneed}, which is usually retrieved by using BM25, dense retrieval model or search engine~\cite{lee2021learning,asai2023self}.
Recently, given the increasing capabilities of LLMs, there is an emerging way to simply concatenate all retrieved data into the context of LLMs for generation~\cite{sun2024block}.

\paragraph{Multi-modal RAG}
Despite the great potential of RAG, it can only handle textual inputs and cannot utilize the rich information in multi-modal data such as images and videos~\cite{su2023pandagpt}.
While earlier efforts to incorporate multi-modal inputs for text generation often relied on carefully designed frameworks~\cite{xiao2024diusum}, a new paradigm has emerged that emphasizes pre-training MLLMs to accomplish this task more directly and efficiently~\cite{chen2024internvl,Qwen2VL,yao2024minicpm,shen2023hugginggpt,riedler2024beyond}, with representative models including GPT-4o~\cite{hurst2024gpt} and Llama-3.2-Vision~\cite{grattafiori2024llama}.
While these multi-modal models greatly expand the range of LLMs' applications, they remain limited to textual outputs, restricting their ability to deliver rich multi-modal responses to users~\cite{zhu2024murar}.

\paragraph{Multi-modal Generation}
In real-world scenarios, humans naturally interact with multi-modal data, such as browsing web pages that combine text, images, and videos in mixed layouts~\cite{zhu2024murar}. Consequently, it is crucial to develop foundation models that not only generate plain text responses to user queries but also incorporate relevant multi-modal data to enhance readability and user engagement. This approach embodies the principle\textemdash\textit{A single image is worth a thousand words}, emphasizing the value of visual elements in effective communication.
To the best of our knowledge, MExBERT~\cite{singh-etal-2021-mimoqa} took the first step in this research direction, which retrieves one image given the user query and generated response. 
Beyond retrieving the images, recent works also focus on generating the corresponding images for model-generated responses, like Next-GPT~\cite{wu2023next}, TextBind~\cite{li2024textbind}, Janus~\cite{wu2024janus} and Emu3~\cite{wang2024emu3}.
Unlike these works, our study focuses on M$^2$RAG task, which aims to dynamically select the multi-modal content from multiple multi-modal inputs (text or images) to construct final multi-modal responses without generating any visual elements.
To the best of our knowledge, MuRAR~\cite{zhu2024murar} is the most closely related works to our study. However, they falls short of modeling multi-modal content holistically. For example, MuRAR generates initial responses as plain text based solely on user queries and retrieved textual content, without incorporating the corresponding multi-modal data. These approaches differs significantly from the setup of our proposed M$^2$RAG task, which emphasizes the understanding the content and relationship among multi-modal input data.

\section{Task Formulation\label{sec:task_formulation}}

In M$^2$RAG task, models generate a multi-modal response $r$ to each user query $Q$ by summarizing a retrieved multi-modal knowledge base $K = \{D_1, \cdots, D_n\}$, consisting of $n$ multi-modal documents or web pages from the Internet.
Each document $D_i$ consists of $m$ ordered elements $\{E_{i,1}, E_{i,2}, \cdots, E_{i,m}\}$, where each element $E_{i,j}$ can be either a text paragraph or an image.
The original order of elements within a document is preserved to ensure images remain paired with their associated textual context, providing rich descriptions that enhance image comprehension.
The process of generating the multi-modal response involves two key steps: \textbf{\textit{In-Doc Retrieval}} and \textbf{\textit{Generation}}.

\paragraph{In-Doc Retrieval}

\begin{figure*}[htbp]
    \centering
    \includegraphics[width=1.0\linewidth]{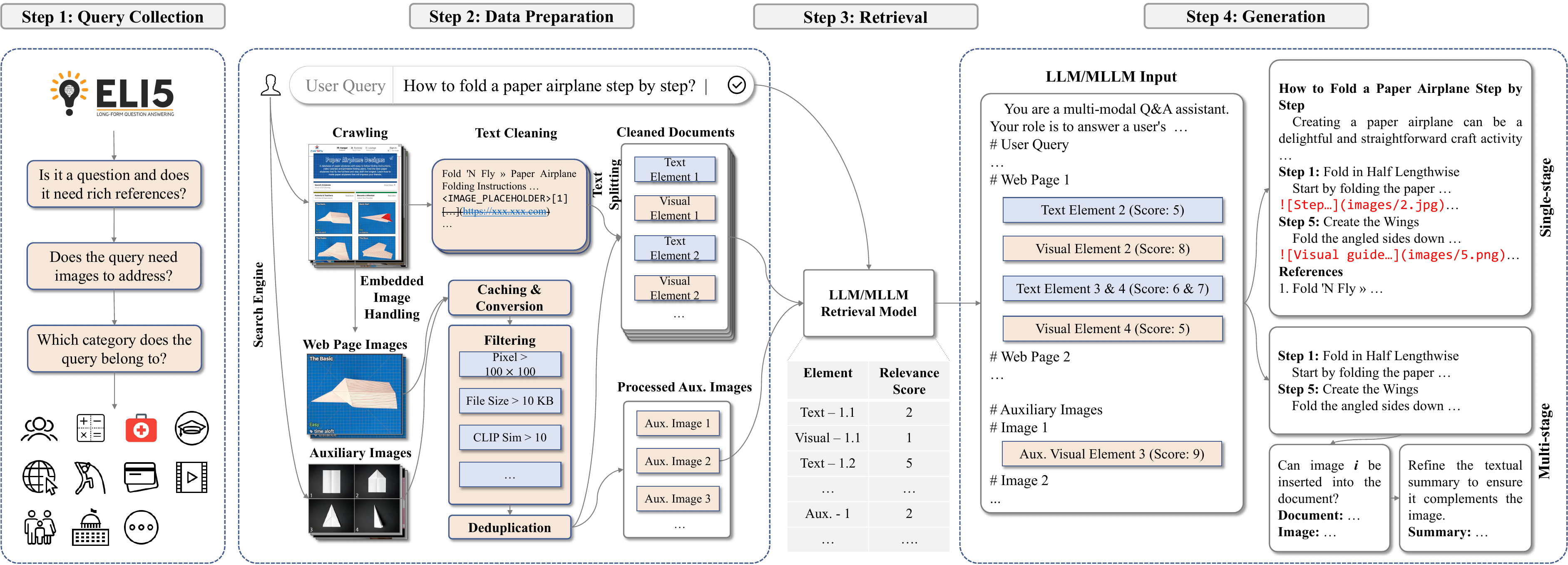}
    \caption{The framework of our proposed dataset construction and M$^2$RAG pipeline. Step 1-3 represent the data curation pipeline and Step 4 demonstrates our proposed generation strategies.}
    \label{fig:framework}
\end{figure*}

While previous studies have demonstrated the strong capabilities of LLMs to answer user queries grounded in retrieved documents, their inference time grows significantly with longer input sequences~\cite{sun2024block}.
This challenge is more severe in the M$^2$RAG task, where models must handle extensive textual and visual data across multiple documents.
To address this challenge, we introduce the In-Doc Retrieval, which selects the most relevant and useful elements from $K$ to reduce inference costs.
As illustrated in Figure~\ref{fig:framework} (Step 3), a retrieval model $M_R$ assesses the relevance of each element $E_{i,j}$ in the knowledge base $K$ with respect to the user query $Q$: $M_R(Q, K)$. The top-$k$ most relevant elements are selected to form a refined and concise knowledge base, $K_{\text{In-Doc}}$. Importantly, when a visual element is selected, its associated textual context is also retrieved to ensure coherence and provide richer input information for the generation process.
The implementation details of the retrieval model is provided in Section~\ref{subsec:element_evaluation}.

\paragraph{Generation} The refined knowledge base $K_{\text{In-Doc}}$ is then processed by the generative model $M_G$ alongside the user query $Q$ to generate the final multi-modal response:
\begin{equation}
    r = M_G(Q, K_{\text{In-Doc}})
\end{equation}
Here, $r$ represents the final answer to the user query. 
Notably, if no visual elements are in $K_{\text{In-Doc}}$, the task defaults to the traditional RAG, since no multi-modal elements are utilized and generated.

\section{Datasets and Methodology for M$^2$RAG~\label{sec:method}}
We describe details of our work:
(1) Benchmark (Section~\ref{subsec:benchmark_preparation}); (2) Evaluation Metrics (Section~\ref{subsec:evalation_metrics}); (3) Generation Strategy (Section~\ref{subsec:generation_strategy}); and (4) Training Dataset (Section~\ref{subsec:training_dataset}).

\subsection{Benchmark Construction\label{subsec:benchmark_preparation}}
Our benchmark are established by running following steps:
(1) Query Collection; (2) Data Preparation; and (3) Retrieval.

\subsubsection{Query Collection}
The queries are sourced from the ELI5~\cite{fan2019eli5} dataset and undergo a series of filtering and classification procedures, as outlined in Step 1 of Figure~\ref{fig:framework}. Specifically, we first exclude queries that do not require rich references and then discard those that do not necessitate visual information for answers.
The filtered queries are then classified into 11 topics to ensure dataset balance and support further investigation of the proposed methods' topic sensitivity.\footnote{See Appendix~\ref{app:query_collection} for details.}

\subsubsection{Data Preparation~\label{subsubsec:data_preparation}}
Subsequently, we crawl the multi-modal data related to each query and post-process them by conducting following steps (Step 2 of Figure~\ref{fig:framework}): (1) Web Page and Auxiliary Image Crawling; (2) Text Processing; and (3) Image Processing. 

\paragraph{Web Page and Auxiliary Image Crawling}
For each collected query, we first employ Google Custom Search API\footnote{\url{https://developers.google.com/custom-search}} to get the URLs of web pages given the query, and then retrieve each web page containing a mix of text and images in markdown format using JINA AI Reader\footnote{\url{https://jina.ai/reader/}}.
Our preliminary study reveals that the images within these web pages are sometimes inaccessible, of low quality, or insufficient for effective multi-modal generation. 
Therefore, we supplement the dataset with auxiliary images sourced from Google Image Search, using the user queries as search input to ensure relevance and enrich the visual content.

\paragraph{Text Processing} 
Raw web pages often include irrelevant or redundant text, such as advertisements and web links, which must be removed to produce concise and relevant content for generation. The text processing pipeline consists of three steps: (1) Embedded Image Handling: Extract image URLs from web pages for image collection and replace them with placeholders in the text; (2) Text Cleaning: Remove text snippets matching predefined patterns, such as web links, to ensure cleaner and more relevant content; (3) Text Segmentation: Employ the text splitter to split the cleaned text into smaller pieces, enabling efficient selection and processing.

\paragraph{Image Processing}
We conduct following steps to process images to ensure they are of high quality and relevant to the user query: 
(1) Caching \& Conversion: 
All images are downloaded using their URLs and converted into widely accepted formats such as JPG, PNG, GIF, or WEBP. Images that cannot be successfully downloaded or converted are discarded; 
(2) Filtering: images smaller than a certain threshold or with a low CLIP-based~\cite{radford2021learning} similarity score to the query text are removed.
Such images often consist of non-representative visual contents, such as  icons, banners, etc.
(3) Deduplication: duplicate or highly similar images are removed using PHash~\cite{zauner2010implementation} algorithm.

\subsubsection{Element Evaluation\label{subsec:element_evaluation}}
As described in Section~\ref{sec:task_formulation}, we need to perform In-Doc retrieval to select top-$k$ relevant text and image elements for efficient generation. 
To achieve this goal, we utilize DeepSeek-Chat~\cite{liu2024deepseek} and MiniCPM-V-2.6~\cite{yao2024minicpm} for evaluation, as shown in Figure~\ref{fig:framework} (Step 3).
Our preliminary manual annotations reveal that these methods demonstrate strong correlations with human evaluations in measuring correlation and significantly outperforms the traditional embedding-based models like CLIP~\cite{radford2021learning}, thereby resulting in better quality of our benchmark\footnote{Please refer to Appendix~\ref{app:element_relecance_eval} for more details.}. 

In summary, given these data resources, we sample 100 queries for each category (other than \textit{Others}), resulting in 1,000 queries for the benchmark dataset.

\subsection{Evaluation Metrics}
\label{subsec:evalation_metrics}
To examine the performance of LLMs and MLLMs on our proposed benchmark, we introduce four text-modal and four multi-modal metrics. 
Since the generated responses are open-ended, most evaluation metrics are implemented by prompting advanced GPT-4o model in a reference-free manner~\cite{lan2024criticeval}.

\paragraph{Text-modal Metrics}
Following prior works~\cite{zhu2024murar}, we evaluate the quality of textual generations using both linguistic and RAG metrics~\cite{es2024ragas,tian-etal-2024-sheng}:
(1) \textbf{Fluency} assesses the linguistic quality of model-generated text, ensuring the grammatical correctness, coherence, and readability~\cite{su2022contrastive,es2024ragas};
(2) \textbf{Relevance} evaluates how well the model-generated textual content aligns with the given user query~\cite{lan2022momentum,es2024ragas};
(3) \textbf{Context Precision} measures the proportion of relevant chunks in $K_{\text{In-Doc}}$ by analyzing the overlap of their key concepts~\cite{es2024ragas}.
(4) \textbf{Faithfulness} measures the accuracy of generations in representing information from $K_{\text{In-Doc}}$, focusing on factual alignment to avoid fabricated or misleading details~\cite{es2024ragas,sun2024block}.

\paragraph{Multi-modal Metrics}
Unlike evaluation metrics that focus solely on image quality~\cite{tu2024automatic}, our work evaluates the interplay between text and images, covering four key aspects:
(1) \textbf{Image Coherence} examines the logical and coherent alignment of images with their surrounding text, ensuring that the visual content enhances and complements the textual narrative~\cite{zhu2024murar};
(2) \textbf{Image Helpfulness} evaluates the contributions of images to the user's understanding of the text, assessing whether visuals provide additional insights, clarify complex ideas, or support textual details~\cite{zhu2024murar};
(3) \textbf{Image Reference} verifies the appropriateness and alignment between images and their textual references. Inadequate or incorrect references of images lead to the confusion during reading;
(4) \textbf{Image Recall} measures the proportion of highly relevant, informative, and important images incorporated into the generations. These crucial images are evaluated and collected by Element Evaluation (Section~\ref{subsec:element_evaluation}).

\paragraph{Overall Score Computation}
We compute the overall score by taking the average of all text-modal and multi-modal metrics, reflecting the overall capabilities of models. Scores are scaled to [0, 100], with higher values indicating superior performance. For additional details on the specific prompts used for each metric, please refer to Appendix~\ref{app:eval_metrics}.

\subsection{Generation Strategy for M$^2$RAG\label{subsec:generation_strategy}}
Previous work MuRAR~\cite{zhu2024murar} generates and integrates text and images independently. We refer to this approach as the separate strategy, which neglects interactions between multi-modal data during text generation.
To address this limitation, we propose two joint modeling strategies that explicitly capture the relationships between text and images: the single-stage strategy and the multi-stage strategy, as illustrated in Step 4 of Figure~\ref{fig:framework}.

\paragraph{Single-stage} requires the model to directly generate a multi-modal output by placing selected images into their corresponding placeholders, using all multi-modal content provided within a single prompt.

\paragraph{Multi-stage} addresses the common limitation of foundation models, which often struggle to process a large number of images simultaneously. It involves three stages: (1) Text Generation: generating a plain text answer based on the multi-modal input; (2) Image Interleaving: dividing the text into segments and prompting LLMs or MLLMs to identify which segments require image insertions for improved readability; (3) Text Refinement: refining each segment by incorporating information from the selected images. 

Both LLMs and MLLMs are able to handle M$^2$RAG task with these two strategies. 
Specifically, LLMs and MLLMs get the prompt $P$ as input to handle M$^2$RAG task, which is constructed using a structured template $T$:
\begin{equation}
    P = T(G, Q, K_{\text{In-Doc}})
\end{equation}
where $G,Q,K_{\text{In-Doc}}$ denotes the task guidelines, user query and knowledge base. 
In the prompt $P$, each image is represented using Markdown format: \verb|![IMAGE_CONTENT](PSEUDO_URL)|: 
(1) \textbf{For MLLMs}, \verb|IMAGE_CONTENT| refers to encoded embeddings of the image, with the entire placeholder positioned where the image originally appears in the context~\cite{grattafiori2024llama,Qwen2VL}, ensuring coherence between the image and the surrounding text;
(2) \textbf{For LLMs}, we convert the image into a detailed textual description as \verb|IMAGE_CONTENT|, enabling the LLMs to comprehend the semantic information of the image. \verb|PSEUDO_URL| serves as the identifier for each input image. In the output of the single-stage approach, images are also represented using the aforementioned Markdown format, where \verb|PSEUDO_URL| indicates the index of the selected image.

\subsection{Training Dataset Construction\label{subsec:training_dataset}}
As another contribution to the research community, we construct a high-quality training dataset to enhance the performance of small-scale but efficient 7B-8B LLMs and MLLMs on the M$^2$RAG task.
Specifically, we first utilize the state-of-the-art GPT-4o with multi-stage strategy to construct 3K triplets of ($Q$, $K_\text{In-Doc}$, $r$).\footnote{Experimental results in Section~\ref{sec:exp_overall_comp} suggest this setup leads to the best performance.}
Then, our proposed multi-modal metrics are used to filter the high-quality samples. 
We do not utilize the text-modal metrics, since the time cost of Context Precision and Faithfulness metrics on 3K long generations are huge\footnote{Please refer to Section~\ref{sec:limitations} for more explanation.}, and we plan to improve the efficiency of text-modal evaluation metrics and update the training data in future work.
Finally, the dataset consists of 1.6K instances.
More implementation details and statistical information of our training dataset could be found in Appendix~\ref{app:dataset_statistics}.

\section{Experimental Results~\label{sec:experiment}}

We conduct the comprehensive evaluation results to demonstrate:
(1) Reliability of our designed multi-modal automatic evaluation metrics (Section~\ref{sec:exp_validate});
(2) A landscape of advanced LLMs and MLLMs performance (Section~\ref{sec:exp_overall_comp});
and (3) Effectiveness of our training dataset (Section~\ref{sec:exp_fine_tuning}).

\subsection{Validate Reliability of Evaluation Metrics}
\label{sec:exp_validate}

\begin{table}[h]
    \centering
    \caption{\label{tab:annotation_correlation}The Spearman's correlation between MLLMs and human annotators. \textbf{Max}, \textbf{Min} and \textbf{Avg.} represents the maximum, minimum and average correlations with all human annotators except for themselves. All evaluations using MLLM are repeated three times, with the average value and standard deviation reported.}
    \resizebox{0.6\linewidth}{!}{
        \begin{tabular}{lllll}
        \toprule
            \textbf{Metric} & \textbf{Human v.s.} & \textbf{Max} & \textbf{Min} & \textbf{Avg.} \\
        \midrule
            \multirow{2}{*}[-0.0ex]{\textbf{Image Coherence}} & \textbf{MLLM} & 
            \textbf{0.58}$_\text{0.023}$ & 0.44$_\text{0.052}$ & \textbf{0.52}$_\text{0.036}$ \\
            & \textbf{Human} & 0.55 & \textbf{0.46} & 0.51 \\
        \midrule
            \multirow{2}{*}[-0.0ex]{\textbf{Image Helpfulness}} & \textbf{MLLM} & \textbf{0.63}$_\text{0.032}$ & 0.45$_\text{0.031}$ & \textbf{0.56}$_\text{0.021}$ \\
            & \textbf{Human} & 0.61 & \textbf{0.47} & 0.53 \\
        \midrule
            \multirow{2}{*}[-0.0ex]{\textbf{Image Reference}} & \textbf{MLLM} & \textbf{0.67}$_\text{0.040}$ & 0.48$_\text{0.031}$ & \textbf{0.59}$_\text{0.038}$ \\
            & \textbf{Human} & \textbf{0.67} & \textbf{0.54} & 0.59 \\
        \bottomrule
        \end{tabular}
    }
\end{table}

Since the reliability of text-modal metrics have been proven in RAG scenarios~\cite{es2024ragas}, we mainly focus on validating the reliability of our proposed MLLM-based multi-modal metrics: (1) Image Coherence; (2) Image Helpfulness; and (3) Image Reference. In practice, we randomly select 200 samples from the benchmark labeled by each metric, and three annotators independently scored these samples using the same scoring criteria as the model. 
Specifically, we calculate the Spearman correlation coefficients between two scorers, each can be human annotators or MLLMs.
Table~\ref{tab:annotation_correlation} reveals that MLLMs achieve comparable performance to inner-correlation of human annotators.
GPT-4o is even better than human average on helpfulness metric (0.57 > 0.53).
These results indicate that our proposed multi-modal evaluation metrics are good proxy for human annotators.

\subsection{Overall Experimental Results\label{sec:exp_overall_comp}}

\begin{table*}[t]
    \tiny
    \centering
    \caption{\label{tab:evaluation_results}The overall experiment results on M$^2$RAG task. \textbf{Flu., Rel., CP. and Faith.} represent fluency, relevance, context precision and faithfulness, respectively. \textbf{Coher., Help., Ref., and Recall} represent image coherence, helpfulness, reference and recall, respectively. The highest scores for each group are highlighted in \textbf{bold}, and the highest scores for all settings are highlighted in \textbf{\textcolor{red}{red}}.}
    \vspace{8.0pt}
    \resizebox{1.0\linewidth}{!}{
        \begin{tabular*}{0.992\linewidth}{cclccccccccc}
        \toprule
            \multirow{2}{*}[-1.0ex]{\textbf{\begin{tabular}[c]{@{}c@{}}Model\\Type\end{tabular}}} & \multirow{2}{*}[-1.0ex]{\textbf{\begin{tabular}[c]{@{}c@{}}Generation\\Strategy\end{tabular}}} & \multirow{2}{*}[-1.0ex]{\textbf{Model}} & \multicolumn{4}{c}{\textbf{Text-modal Metrics}} & \multicolumn{4}{c}{\textbf{Multi-modal Metrics}} & \multirow{2}{*}[-1.0ex]{\textbf{Overall}} \\
        \cmidrule{4-11}
            & & & \textbf{Flu.} & \textbf{Rel.} & \textbf{CP.} & \textbf{Faith.} & \textbf{Coher.} & \textbf{Help.} & \textbf{Ref.} & \textbf{Recall} & \\
        \midrule
            \multirow{4}{*}[-1.0ex]{\textbf{Reasoners}} & \multirow{2}{*}{\textbf{Single}} & OpenAI o3-mini & 80.8 & \textcolor{red}{\textbf{89.5}} & 91.9 & \textbf{90.0} & 73.3 & \textbf{63.2} & \textbf{64.6} & 79.1 & \textbf{79.1} \\
            & & DeepSeek-R1 & \textcolor{red}{\textbf{87.4}} & 87.6 & \textbf{92.3} & 85.2 & \textcolor{red}{\textbf{77.9}} & 62.0 & 50.4 & \textbf{85.2} & 78.5 \\
        \cmidrule{2-12}
            & \multirow{2}{*}{\textbf{Multi}} & OpenAI o3-mini & 79.6 & 80.2 & \textbf{94.4} & \textbf{83.8} & 69.7 & 60.6 & 78.0 & \textcolor{red}{\textbf{100.0}} & 80.8 \\
            & & DeepSeek-R1 & \textbf{84.2} & \textbf{84.5} & 83.8 & 75.6 & \textbf{77.1} & \textcolor{red}{\textbf{69.7}} & \textbf{80.8} & \textcolor{red}{\textbf{100.0}} & \textbf{81.9} \\
        \midrule
            \multirow{15}{*}[-4.0ex]{\textbf{LLMs}} & \textbf{Separate} & GPT-4o~\cite{zhu2024murar} & 84.5 & 88.6 & 95.0 & 89.6 & 58.3 & 50.4 & 43.8 & 82.8 & 74.1 \\
        \cmidrule{2-12}
            & \multirow{7}{*}{\textbf{Single}} & GPT-4o & \textbf{82.9} & 88.3 & 95.7 & \textbf{88.8} & \textbf{72.2} & \textbf{63.8} & 32.2 & 75.8 & \textbf{74.9} \\
            & & DeepSeek-V3 & 81.2 & 87.9 & 91.9 & 87.4 & 63.6 & 52.7 & \textbf{36.8} & \textbf{87.3} & 73.6 \\
            & & Llama-3.1-70B-Instruct & 81.2 & 86.5 & 93.6 & 83.8 & 62.7 & 55.9 & 23.8 & 66.0 & 69.2 \\
            & & Qwen2.5-72B-Instruct & 82.5 & \textbf{89.2} & \textbf{95.8} & 87.2 & 66.3 & 59.6 & 36.6 & 78.6 & 74.5 \\
            & & Llama-3.1-8B-Instruct & 79.8 & 85.2 & 91.0 & 81.1 & 54.1 & 46.6 & 26.3 & 79.5 & 68.0 \\
            & & Qwen2.5-7B-Instruct & 74.2 & 79.4 & 88.9 & 82.1 & 55.9 & 48.4 & 22.2 & 80.0 & 66.4 \\
        \cmidrule{3-12}
            & & \cellcolor{gray!20} \textbf{Average} & \cellcolor{gray!20} 80.3 & \cellcolor{gray!20} 86.1 & \cellcolor{gray!20} 92.8 & \cellcolor{gray!20} 85.1 & \cellcolor{gray!20} 62.5 & \cellcolor{gray!20} 54.5 & \cellcolor{gray!20} 29.6 & \cellcolor{gray!20} 77.8 & \cellcolor{gray!20} 71.1 \\
        \cmidrule{2-12}
            & \multirow{7}{*}{\textbf{Multi}} & GPT-4o & 82.5 & 87.5 & 96.6 & 81.8 & 73.3 & \textbf{65.8} & \textbf{81.5} & \textcolor{red}{\textbf{100.0}} & 83.6 \\
            & & DeepSeek-V3 & 80.5 & 83.0 & 95.0 & \textbf{85.9} & 70.0 & 63.1 & 79.1 & \textcolor{red}{\textbf{100.0}} & 82.1 \\
            & & Llama-3.1-70B-Instruct & 80.7 & 87.6 & 97.3 & 83.9 & 73.0 & 63.6 & 76.8 & 99.2 & 82.8 \\
            & & Qwen2.5-72B-Instruct & \textbf{83.8} & \textbf{88.6} & \textcolor{red}{\textbf{98.4}} & 83.1 & \textbf{73.4} & 64.5 & 78.5 & 99.9 & \textcolor{red}{\textbf{83.8}} \\
            & & Llama-3.1-8B-Instruct & 81.3 & 86.5 & 95.8 & 82.1 & 72.1 & 62.0 & 75.1 & 97.9 & 81.6 \\
            & & Qwen2.5-7B-Instruct & 82.8 & 87.0 & 95.3 & 78.5 & 71.3 & 62.4 & 77.9 & 99.0 & 81.8 \\
        \cmidrule{3-12}
            & & \cellcolor{gray!20} \textbf{Average} & \cellcolor{gray!20} 81.9 & \cellcolor{gray!20} 86.7 & \cellcolor{gray!20} 96.4 & \cellcolor{gray!20} 82.5 & \cellcolor{gray!20} 72.2 & \cellcolor{gray!20} 63.6 & \cellcolor{gray!20} 78.1 & \cellcolor{gray!20} 99.3 & \cellcolor{gray!20} 82.6 \\
        \midrule
            \multirow{14}{*}[-3.0ex]{\textbf{MLLMs}} & \multirow{7}{*}{\textbf{Single}} & GPT-4o & 81.4 & \textbf{88.6} & \textbf{95.2} & 88.0 & \textbf{69.8} & \textbf{61.0} & 25.1 & 81.5 & 73.8 \\
            & & Step-1o & \textbf{84.1} & 87.4 & 94.9 & \textcolor{red}{\textbf{90.7}} & 65.9 & 53.8 & \textbf{44.6} & \textbf{84.7} & \textbf{75.8} \\
            & & Llama-3.2-90B-V-Instruct & 82.5 & 73.9 & 77.4 & 79.0 & 55.7 & 51.4 & 18.6 & 58.0 & 62.1 \\
            & & Qwen2-VL-72B-Instruct & 82.5 & 85.3 & 88.3 & 87.9 & 50.9 & 42.1 & 18.8 & 62.6 & 64.8 \\
            & & Llama-3.2-11B-V-Instruct & 83.0 & 71.4 & 73.0 & 70.7 & 37.8 & 20.6 & 1.1 & 58.0 & 51.9 \\
            & & Qwen2-VL-7B-Instruct & 69.0 & 70.6 & 79.0 & 72.9 & 44.7 & 39.3 & 23.6 & 61.5 & 57.6 \\
        \cmidrule{3-12}
            & & \cellcolor{gray!20} \textbf{Average} & \cellcolor{gray!20} 80.4 & \cellcolor{gray!20} 79.5 & \cellcolor{gray!20} 84.6 & \cellcolor{gray!20} 81.5 & \cellcolor{gray!20} 54.1 & \cellcolor{gray!20} 44.7 & \cellcolor{gray!20} 22.0 & \cellcolor{gray!20} 67.7 & \cellcolor{gray!20} 64.3 \\
        \cmidrule{2-12}
            & \multirow{7}{*}{\textbf{Multi}} & GPT-4o & 81.6 & \textbf{88.2} & 96.6 & 81.0 & \textbf{72.2} & 63.5 & 80.6 & \textcolor{red}{\textbf{100.0}} & \textbf{83.0} \\
            & & Step-1o & 80.5 & 82.5 & \textbf{96.7} & \textbf{89.8} & 71.6 & \textbf{63.7} & \textcolor{red}{\textbf{81.6}} & 97.6 & \textbf{83.0} \\
            & & Llama-3.2-90B-V-Instruct & 80.9 & 82.8 & 92.0 & 75.5 & 60.5 & 51.0 & 62.7 & 98.1 & 75.4 \\
            & & Qwen2-VL-72B-Instruct & \textbf{83.0} & 86.5 & 94.3 & 79.0 & 68.0 & 58.8 & 71.2 & 98.8 & 79.9 \\
            & & Llama-3.2-11B-V-Instruct & 82.8 & 80.7 & 87.8 & 69.9 & 52.7 & 42.7 & 34.4 & 86.2 & 67.1 \\
            & & Qwen2-VL-7B-Instruct & 80.6 & 86.7 & 93.1 & 79.2 & 60.3 & 49.1 & 49.9 & 96.0 & 74.4 \\
        \cmidrule{3-12}
            & & \cellcolor{gray!20} \textbf{Average} & \cellcolor{gray!20} 81.6 & \cellcolor{gray!20} 84.5 & \cellcolor{gray!20} 93.4 & \cellcolor{gray!20} 79.1 & \cellcolor{gray!20} 64.2 & \cellcolor{gray!20} 54.8 & \cellcolor{gray!20} 63.4 & \cellcolor{gray!20} 96.1 & \cellcolor{gray!20} 77.1 \\
        \midrule
            \multirow{3}{*}[-0.0ex]{\textbf{\begin{tabular}[c]{@{}c@{}}Fine-tuned\\Models\end{tabular}}} & \multirow{3}{*}[-0.0ex]{\textbf{Single}} & Llama-3.1-8B-Instruct & 74.9 & 76.5 & 96.1 & 82.4 & 66.3 & 60.6 & 75.2 & 97.0 & 78.6 \\
            & & Qwen2.5-7B-Instruct & 75.3 & 77.6 & 94.9 & 80.6 & 66.0 & 58.8 & 73.9 & 87.6 & 76.9 \\
            & & Qwen2-VL-7B-Instruct & 57.2 & 74.3 & 91.4 & 71.4 & 65.4 & 57.4 & 73.5 & 95.0 & 73.2 \\
        \bottomrule
        \end{tabular*}
    }
\end{table*}

We conduct comprehensive experiments on our benchmark with 12 representative models, including OpenAI o3-mini, DeepSeek-R1, GPT-4o, DeepSeek-V3, Step-1o, Llama-3.1 \& Llama-3.2-Vision series, Qwen2.5 \& Qwen2-VL series (see Appendix~\ref{app:experimental_setup} for details). Due to the high evaluation cost, we sample 200 queries from the benchmark dataset by category to compare the performance of a wider range of models. Meanwhile, we also conduct experiments on the entire dataset containing 1,000 queries to further validate our findings.
The results are shown in Table~\ref{tab:evaluation_results} and Table~\ref{tab:eval_results_entire_dataset}, and we can observe several findings.

\paragraph{The Most Capable Models}
Table~\ref{tab:evaluation_results} indicates that the most capable open-source and closed-source LLMs for the M$^2$RAG task are Qwen2.5-72B-Instruct and GPT-4o, using a multi-stage strategy with LLM backbones. 
Among MLLM backbones, GPT-4o and Step-1o outperform all open-source models. For example, Step-1o surpassing Qwen2-VL-72B-Instruct by a large margin (83.0 > 79.9). 
The reasoners OpenAI o3-mini and DeepSeek-R1 significantly outperform other models in the single-stage approach.
Despite these strong performances, there remains significant room for improvement, as even the best-performing model falls short of the ideal score (83.8 < 100.0). 

\paragraph{Reasoners Excel in Single-stage Generation}
In single-stage settings, reasoner models (e.g., OpenAI o3-mini and DeepSeek-R1) significantly outperform standard LLMs, owing to their inherent chain-of-thought reasoning that guides both text and image selection. However, this advantage narrows under the multi-stage strategy: the explicit image-interleaving stages compensate for standard LLMs’ lack of built-in reasoning, allowing them to close the gap.

\paragraph{Abnormal Performance of Open-source MLLMs Reflect Pretraining Differences}
The MLLMs with comparable capabilities, Llama-3.2-Vision and Qwen2-VL series, show great performance gap in the M$^2$RAG task. The poor performance pf Llama-3.2-Vision is likely due to its pretraining on single image-text pairs~\cite{grattafiori2024llama}, which limits its ability to handle multiple images. Qwen2-VL, on the other hand, has multi-image pretraining and thus supports multi-image reasoning and demonstrates more coherent output in multi-modal RAG tasks~\cite{Qwen2VL}.

\paragraph{LLMs generally outperform open-source MLLMs of similar size}
Due to the multi-image confusion phenomenon, current MLLMs struggle with reasoning over multiple images, aligning with the findings of previous works~\cite{liu2024mibench}. In contrast, LLMs receive detailed image descriptions and do not directly process raw visual input, avoiding this issue.

\paragraph{Scaling Phenomena} A clear scaling trend is observed, where larger open-source models generally yield better results, except in the single-stage approach with LLMs, where Llama-3.1-8B-Instruct surprisingly outperforms Llama-3.1-70B-Instruct in Image Recall (79.5 > 66.0). This discrepancy arises because larger models tend to select fewer images, some of which are beneficial, leading to degraded performance. 

\paragraph{Comparison between Separate and Joint Modeling} Regarding modeling approaches, separate modeling~\cite{zhu2024murar} proves significantly weaker than joint modeling, with even the worst-performing joint approach (GPT-4o, 74.9) surpassing separate modeling (74.1), highlighting the importance of modeling multi-modal interactions.

\paragraph{Comparison among Strategies} Multi-stage approaches consistently outperform single-stage methods for both LLMs (82.6 > 71.1) and MLLMs (77.1 > 64.3), as they introduce more relevant images, improving information density and readability. Additionally, open-source LLMs significantly outperform MLLMs across both modeling paradigms, with an overall score advantage (76.8 > 70.7), indicating that current open-source MLLMs still struggle with the complexity of M$^2$RAG tasks.

\subsection{Improvements from Fine-tuning}
\label{sec:exp_fine_tuning}

We further conduct the supervised fine-tuning to improve the 7B-8B LLMs and MLLMs capabilities in M$^2$RAG task. Please refer to Appendix~\ref{app:sft_settings} for more fine-tuning implementation details. The evaluation results are illustrated in Table~\ref{tab:evaluation_results}. It can be observed that our fine-tuned LLMs and MLLMs exhibit significant improvements on all multi-modal evaluation metrics and the overall score.
For example, the average improvements on the overall score is 19.2\%.
Since the text-modal metrics are not utilized for constructing training dataset, our fine-tuned models underperforms the baselines on some textual evaluation metrics, particularly Fluency.
Though the text-modal metrics are not our primary focus, it is still worth exploring whether the involving of these metrics in training data filtering can further improve the overall performance of open-source LLMs and MLLMs.
\section{Analysis}

In this section, we present more analysis on our benchmark by answering several questions.

\subsection{Does Topic Affect Performance?}

\begin{wrapfigure}{r}{0.4\textwidth}
    \vspace{-10pt}
    \resizebox{\linewidth}{!}{%
        \includegraphics[width=\textwidth]{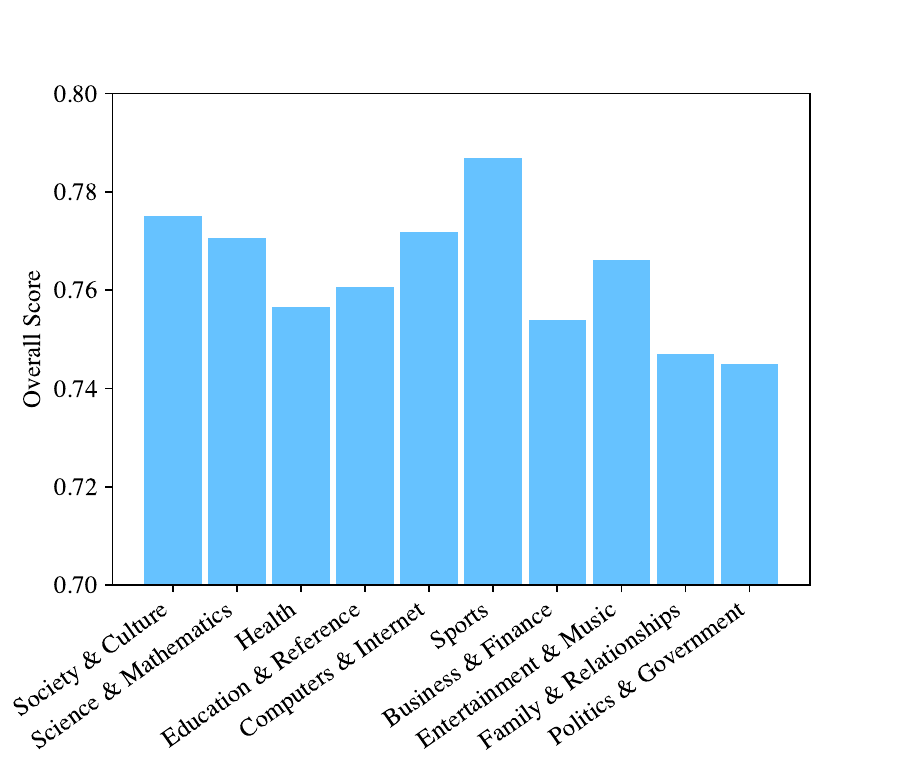}
    }
    \caption{Average overall score across 10 topics.}
    \label{fig:category_scores}
    \vspace{-20pt}
\end{wrapfigure}

The average overall scores of all evaluated models, as presented in Figure~\ref{fig:category_scores}, reveal notable variations in performance across different topics. For instance, the most challenging topic is \textit{\textbf{Politics \& Government}}, which achieves the lowest overall score, whereas \textit{\textbf{Society \& Culture}} and \textit{\textbf{Sports}} are significantly easier topics by comparison.

We hypothesize that these performance differences stem from the varying demands for multi-modal information across topics. Specifically, images are more prevalent and useful in less technical fields, such as sports and culture. In contrast, fields like politics and finance often require more accurate data from additional modalities, such as tables and charts, which are less readily integrated. 
These phenomena suggest that future work needs to carefully implement strategies to control the proportion of included image across different topics.

\subsection{Is Description Quality Important?}

\begin{wraptable}{r}{0.5\textwidth}
    \centering
    \vspace{-15pt}
    \caption{Results of ablation study on the quality of image descriptions and the inclusion of auxiliary images. \textbf{Text Avg.} and \textbf{Multi-Modal Precision Avg.} represent the average values of text-modal metrics and multi-modal precision metrics, respectively. Recall and Overall scores for \textbf{w/o Aux. Image} is not reported due to changes in the images within the knowledge base.}
    \vspace{5pt}
    \resizebox{1.0\linewidth}{!}{
        \begin{tabular}{ccccc}
            \toprule
                \textbf{Ablation} & \textbf{\begin{tabular}[c]{@{}c@{}}Text\\Avg.\end{tabular}} & \textbf{\begin{tabular}[c]{@{}c@{}}Multi-Modal\\Precision Avg.\end{tabular}} & \textbf{\begin{tabular}[c]{@{}c@{}}Recall\\Score\end{tabular}} & \textbf{\begin{tabular}[c]{@{}c@{}}Overall\\Score\end{tabular}} \\
            \midrule
                \textbf{Original} & \textbf{88.9} & \textbf{56.1} & 75.8 & \textbf{74.9} \\
            \midrule
                \textbf{w/o Context} & 88.5 & 51.4 & \textbf{76.8} & 73.1 \\
                \textbf{w/o Detail} & 88.8 & 54.4 & 73.5 & 74.0 \\
                \textbf{w/o Aux. Image} & 88.7 & 52.4 & - & - \\
            \bottomrule
        \end{tabular}
    }
    \vspace{-10pt}
    \label{tab:ablation}
\end{wraptable}

As described in Section~\ref{subsec:generation_strategy}, each image is transcribed into a detailed textual description with its context for LLMs.
Here we conduct ablation studies on the quality of image description. As illustrated in Table~\ref{tab:ablation}, we can conclude that (1) the exclusion of context in description generation leads to a significant decline in model performance under most metrics; (2) the absence of details in descriptions leads to a notable drop in Image Recall, highlighting the critical role of details in guiding image selection for LLMs. Overall, incorporating context and details in the image descriptions contribute to GPT-4o's performance to a measurable extent.

\subsection{Are Auxiliary Images Helpful?}
As mentioned in Section~\ref{subsubsec:data_preparation}, auxiliary images are collected to address cases where relevant images are absent from the retrieved web pages. 
As illustrated in last row of Table~\ref{tab:ablation} (w/o Aux. Image), removing these auxiliary images leads to a significant decline in generative performance in multi-modal metrics. This observation reflects the challenges in effectively integrating auxiliary images that lack contextual information with the textual content.
Overall, the inclusion of auxiliary images significantly enriches the visual content and enhances the quality of the generated document.
\section{Conclusion}

In this paper, we formate a challenging task\textemdash M$^2$RAG, requiring foundation models to process the mixed multi-modal web pages to generate a multi-modal answers for solving user queries.
Besides, we construct a benchmark to comprehensively evaluate the capabilities of existing foundation models based on four text-only and five multi-modal fine-grained metrics.
Furthermore, we also propose several strong baselines for existing models to solve this task.
Extensive experimental results demonstrate several intriguing phenomena, facilitating the future research in this field.

\bibliographystyle{ref}  
\small
\bibliography{reference}

\begin{thebibliography}{41}
\providecommand{\natexlab}[1]{#1}
\providecommand{\url}[1]{\texttt{#1}}
\expandafter\ifx\csname urlstyle\endcsname\relax
  \providecommand{\doi}[1]{doi: #1}\else
  \providecommand{\doi}{doi: \begingroup \urlstyle{rm}\Url}\fi

\bibitem[Asai et~al.(2023)Asai, Wu, Wang, Sil, and Hajishirzi]{asai2023self}
Asai , A., Wu~, Z., Wang , Y., Sil , A., \& Hajishirzi , H. (2023)
\newblock Self-rag: Learning to retrieve, generate, and critique through self-reflection. In
\newblock \emph{The Twelfth International Conference on Learning Representations}

\bibitem[Chen et~al.(2022)Chen, Hu, Chen, Verga, and Cohen]{chen2022murag}
Chen , W., Hu~, H., Chen , X., Verga , P., \& Cohen , W.~W. (2022)
\newblock Murag: Multimodal retrieval-augmented generator for open question answering over images and text.
\newblock \emph{arXiv preprint arXiv:2210.02928}

\bibitem[Chen et~al.(2024)Chen, Wu, Wang, Su, Chen, Xing, Zhong, Zhang, Zhu, Lu, et~al.]{chen2024internvl}
Chen , Z., Wu~, J., Wang , W., Su~, W., Chen , G., Xing , S., Zhong , M., Zhang , Q., Zhu , X., Lu~, L., \& others  (2024)
\newblock Internvl: Scaling up vision foundation models and aligning for generic visual-linguistic tasks. In
\newblock \emph{Proceedings of the IEEE/CVF conference on computer vision and pattern recognition}
\newblock pages 24185--24198.

\bibitem[Dao(2024)]{dao2023flashattention2}
Dao , T. (2024)
\newblock Flash{A}ttention-2: Faster attention with better parallelism and work partitioning. In
\newblock \emph{International Conference on Learning Representations (ICLR)}

\bibitem[Es et~al.(2024)Es, James, Anke, and Schockaert]{es2024ragas}
Es~, S., James , J., Anke , L.~E., \& Schockaert , S. (2024)
\newblock Ragas: Automated evaluation of retrieval augmented generation. In
\newblock \emph{Proceedings of the 18th Conference of the European Chapter of the Association for Computational Linguistics: System Demonstrations}
\newblock pages 150--158.

\bibitem[Fan et~al.(2019)Fan, Jernite, Perez, Grangier, Weston, and Auli]{fan2019eli5}
Fan , A., Jernite , Y., Perez , E., Grangier , D., Weston , J., \& Auli , M. (2019)
\newblock Eli5: Long form question answering.
\newblock \emph{arXiv preprint arXiv:1907.09190}

\bibitem[Grattafiori et~al.(2024)Grattafiori, Dubey, Jauhri, Pandey, Kadian, Al-Dahle, Letman, Mathur, Schelten, Vaughan, et~al.]{grattafiori2024llama}
Grattafiori , A., Dubey , A., Jauhri , A., Pandey , A., Kadian , A., Al-Dahle , A., Letman , A., Mathur , A., Schelten , A., Vaughan , A., \& others  (2024)
\newblock The llama 3 herd of models.
\newblock \emph{arXiv preprint arXiv:2407.21783}

\bibitem[Hara et~al.(2018)Hara, Adams, Milland, Savage, Callison-Burch, and Bigham]{hara2018data}
Hara , K., Adams , A., Milland , K., Savage , S., Callison-Burch , C., \& Bigham , J.~P. (2018)
\newblock A data-driven analysis of workers' earnings on amazon mechanical turk. In
\newblock \emph{Proceedings of the 2018 CHI conference on human factors in computing systems}
\newblock pages 1--14.

\bibitem[Hsu et~al.(2024)Hsu, Dai, Kothapalli, Song, Tang, Zhu, Shimizu, Sahni, Ning, and Chen]{hsu2024ligerkernelefficienttriton}
Hsu , P.-L., Dai , Y., Kothapalli , V., Song , Q., Tang , S., Zhu , S., Shimizu , S., Sahni , S., Ning , H., \& Chen , Y. (2024)
\newblock Liger kernel: Efficient triton kernels for llm training.
\newblock \emph{arXiv preprint arXiv:2410.10989}

\bibitem[Hu et~al.(2021)Hu, Shen, Wallis, Allen-Zhu, Li, Wang, Wang, and Chen]{hu2021lora}
Hu~, E.~J., Shen , Y., Wallis , P., Allen-Zhu , Z., Li~, Y., Wang , S., Wang , L., \& Chen , W. (2021)
\newblock Lora: Low-rank adaptation of large language models.
\newblock \emph{arXiv preprint arXiv:2106.09685}

\bibitem[Hurst et~al.(2024)Hurst, Lerer, Goucher, Perelman, Ramesh, Clark, Ostrow, Welihinda, Hayes, Radford, et~al.]{hurst2024gpt}
Hurst , A., Lerer , A., Goucher , A.~P., Perelman , A., Ramesh , A., Clark , A., Ostrow , A., Welihinda , A., Hayes , A., Radford , A., \& others  (2024)
\newblock Gpt-4o system card.
\newblock \emph{arXiv preprint arXiv:2410.21276}

\bibitem[Joshi et~al.(2024)Joshi, Gupta, Kumar, and Sisodia]{joshi2024robust}
Joshi , P., Gupta , A., Kumar , P., \& Sisodia , M. (2024)
\newblock Robust multi model rag pipeline for documents containing text, table \& images. In
\newblock \emph{2024 3rd International Conference on Applied Artificial Intelligence and Computing (ICAAIC)}
\newblock pages 993--999. IEEE.

\bibitem[Kwon et~al.(2023)Kwon, Li, Zhuang, Sheng, Zheng, Yu, Gonzalez, Zhang, and Stoica]{kwon2023efficient}
Kwon , W., Li~, Z., Zhuang , S., Sheng , Y., Zheng , L., Yu~, C.~H., Gonzalez , J.~E., Zhang , H., \& Stoica , I. (2023)
\newblock Efficient memory management for large language model serving with pagedattention. In
\newblock \emph{Proceedings of the ACM SIGOPS 29th Symposium on Operating Systems Principles}

\bibitem[Lan et~al.(2022)Lan, Su, Liu, Huang, and Mao]{lan2022momentum}
Lan , T., Su~, Y., Liu , S., Huang , H., \& Mao , X.-L. (2022)
\newblock Momentum decoding: Open-ended text generation as graph exploration.
\newblock \emph{arXiv preprint arXiv:2212.02175}

\bibitem[Lan et~al.(2023)Lan, Cai, Wang, Huang, and Mao]{lan2023copyneed}
Lan , T., Cai , D., Wang , Y., Huang , H., \& Mao , X.-L. (2023)
\newblock Copy is all you need. In
\newblock \emph{The Eleventh International Conference on Learning Representations}

\bibitem[Lan et~al.(2024{\natexlab{a}})Lan, Ma, Zhou, Xu, and Mao]{tian-etal-2024-sheng}
Lan , T., Ma~, Z.-A., Zhou , Y., Xu~, C., \& Mao , X.-L. (2024.
\newblock {\natexlab{a}}) A survey of automatic evaluation on the quality of generated text. In
\newblock X.~Zhao, editor,
\newblock \emph{Proceedings of the 23rd Chinese National Conference on Computational Linguistics (Volume 2: Frontier Forum)}
\newblock pages 169--196, Taiyuan, China: Chinese Information Processing Society of China.

\bibitem[Lan et~al.(2024{\natexlab{b}})Lan, Zhang, Xu, Huang, Lin, Chen, and Mao]{lan2024criticeval}
Lan , T., Zhang , W., Xu~, C., Huang , H., Lin , D., Chen , K., \& Mao , X.-l. (2024.
\newblock {\natexlab{b}})
\newblock Criticeval: Evaluating large language model as critic.
\newblock \emph{arXiv preprint arXiv:2402.13764}

\bibitem[Lee et~al.(2021)Lee, Sung, Kang, and Chen]{lee2021learning}
Lee , J., Sung , M., Kang , J., \& Chen , D. (2021)
\newblock Learning dense representations of phrases at scale. In
\newblock \emph{Proceedings of the 59th Annual Meeting of the Association for Computational Linguistics and the 11th International Joint Conference on Natural Language Processing (Volume 1: Long Papers)}
\newblock pages 6634--6647.

\bibitem[Li et~al.(2022)Li, Su, Cai, Wang, and Liu]{li2022survey}
Li~, H., Su~, Y., Cai , D., Wang , Y., \& Liu , L. (2022)
\newblock A survey on retrieval-augmented text generation.
\newblock \emph{arXiv preprint arXiv:2202.01110}

\bibitem[Li et~al.(2024)Li, Li, Cai, Wang, Liu, Watanabe, Yang, and Shi]{li2024textbind}
Li~, H., Li~, S., Cai , D., Wang , L., Liu , L., Watanabe , T., Yang , Y., \& Shi , S. (2024)
\newblock Textbind: Multi-turn interleaved multimodal instruction-following in the wild. In
\newblock \emph{Findings of the Association for Computational Linguistics ACL 2024}
\newblock pages 9053--9076.

\bibitem[Liu et~al.(2024{\natexlab{a}})Liu, Feng, Wang, Wang, Liu, Zhao, Dengr, Ruan, Dai, Guo, et~al.]{liu2024deepseek}
Liu , A., Feng , B., Wang , B., Wang , B., Liu , B., Zhao , C., Dengr , C., Ruan , C., Dai , D., Guo , D., \& others  (2024.
\newblock {\natexlab{a}})
\newblock Deepseek-v2: A strong, economical, and efficient mixture-of-experts language model.
\newblock \emph{arXiv preprint arXiv:2405.04434}

\bibitem[Liu et~al.(2024{\natexlab{b}})Liu, Zhang, Xu, Shi, Jiang, Yan, Zhang, Huang, Yuan, Li, et~al.]{liu2024mibench}
Liu , H., Zhang , X., Xu~, H., Shi , Y., Jiang , C., Yan , M., Zhang , J., Huang , F., Yuan , C., Li~, B., \& others  (2024.
\newblock {\natexlab{b}}) Mibench: Evaluating multimodal large language models over multiple images. In
\newblock \emph{Proceedings of the 2024 Conference on Empirical Methods in Natural Language Processing}
\newblock pages 22417--22428.

\bibitem[Radford et~al.(2021)Radford, Kim, Hallacy, Ramesh, Goh, Agarwal, Sastry, Askell, Mishkin, Clark, et~al.]{radford2021learning}
Radford , A., Kim , J.~W., Hallacy , C., Ramesh , A., Goh , G., Agarwal , S., Sastry , G., Askell , A., Mishkin , P., Clark , J., \& others  (2021)
\newblock Learning transferable visual models from natural language supervision. In
\newblock \emph{International conference on machine learning}
\newblock pages 8748--8763. PMLR.

\bibitem[Rajbhandari et~al.(2020)Rajbhandari, Rasley, Ruwase, and He]{rajbhandari2020zero}
Rajbhandari , S., Rasley , J., Ruwase , O., \& He~, Y. (2020)
\newblock Zero: Memory optimizations toward training trillion parameter models. In
\newblock \emph{SC20: International Conference for High Performance Computing, Networking, Storage and Analysis}
\newblock pages 1--16. IEEE.

\bibitem[Riedler and Langer(2024)]{riedler2024beyond}
Riedler , M. \& Langer , S. (2024)
\newblock Beyond text: Optimizing rag with multimodal inputs for industrial applications.
\newblock \emph{arXiv preprint arXiv:2410.21943}

\bibitem[Shen et~al.(2023)Shen, Song, Tan, Li, Lu, and Zhuang]{shen2023hugginggpt}
Shen , Y., Song , K., Tan , X., Li~, D., Lu~, W., \& Zhuang , Y. (2023)
\newblock Hugginggpt: Solving ai tasks with chatgpt and its friends in hugging face.
\newblock \emph{Advances in Neural Information Processing Systems} {\bfseries 36}:\penalty0 38154--38180.

\bibitem[Singh et~al.(2021)Singh, Nasery, Mehta, Agarwal, Lamba, and Srinivasan]{singh-etal-2021-mimoqa}
Singh , H., Nasery , A., Mehta , D., Agarwal , A., Lamba , J., \& Srinivasan , B.~V. (2021)
\newblock {MIMOQA}: Multimodal input multimodal output question answering. In
\newblock K.~Toutanova, A.~Rumshisky, L.~Zettlemoyer, D.~Hakkani-Tur, I.~Beltagy, S.~Bethard, R.~Cotterell, T.~Chakraborty, and Y.~Zhou, (eds.),
\newblock \emph{Proceedings of the 2021 Conference of the North American Chapter of the Association for Computational Linguistics: Human Language Technologies}
\newblock pages 5317--5332, Online: Association for Computational Linguistics.

\bibitem[Su et~al.(2022)Su, Lan, Wang, Yogatama, Kong, and Collier]{su2022contrastive}
Su~, Y., Lan , T., Wang , Y., Yogatama , D., Kong , L., \& Collier , N. (2022)
\newblock A contrastive framework for neural text generation.
\newblock \emph{Advances in Neural Information Processing Systems} {\bfseries 35}:\penalty0 21548--21561.

\bibitem[Su et~al.(2023)Su, Lan, Li, Xu, Wang, and Cai]{su2023pandagpt}
Su~, Y., Lan , T., Li~, H., Xu~, J., Wang , Y., \& Cai , D. (2023)
\newblock Pandagpt: One model to instruction-follow them all.
\newblock \emph{arXiv preprint arXiv:2305.16355}

\bibitem[Sun et~al.(2024)Sun, Wang, and Tian]{sun2024block}
Sun , E., Wang , Y., \& Tian , L. (2024)
\newblock Block-attention for efficient rag.
\newblock \emph{arXiv preprint arXiv:2409.15355}

\bibitem[Tu et~al.(2024)Tu, Ma, Lan, Zhao, Huang, and Mao]{tu2024automatic}
Tu~, R.-C., Ma~, Z.-A., Lan , T., Zhao , Y., Huang , H., \& Mao , X.-L. (2024)
\newblock Automatic evaluation for text-to-image generation: Task-decomposed framework, distilled training, and meta-evaluation benchmark.
\newblock \emph{arXiv preprint arXiv:2411.15488}

\bibitem[Wang et~al.(2024{\natexlab{a}})Wang, Bai, Tan, Wang, Fan, Bai, Chen, Liu, Wang, Ge, Fan, Dang, Du, Ren, Men, Liu, Zhou, Zhou, and Lin]{Qwen2VL}
Wang , P., Bai , S., Tan , S., Wang , S., Fan , Z., Bai , J., Chen , K., Liu , X., Wang , J., Ge~, W., Fan , Y., Dang , K., Du~, M., Ren , X., Men , R., Liu , D., Zhou , C., Zhou , J., \& Lin , J. (2024.
\newblock {\natexlab{a}})
\newblock Qwen2-vl: Enhancing vision-language model's perception of the world at any resolution.
\newblock \emph{arXiv preprint arXiv:2409.12191}

\bibitem[Wang et~al.(2024{\natexlab{b}})Wang, Zhang, Luo, Sun, Cui, Wang, Zhang, Wang, Li, Yu, et~al.]{wang2024emu3}
Wang , X., Zhang , X., Luo , Z., Sun , Q., Cui , Y., Wang , J., Zhang , F., Wang , Y., Li~, Z., Yu~, Q., \& others  (2024.
\newblock {\natexlab{b}})
\newblock Emu3: Next-token prediction is all you need.
\newblock \emph{arXiv preprint arXiv:2409.18869}

\bibitem[Wu et~al.(2024)Wu, Chen, Wu, Ma, Liu, Pan, Liu, Xie, Yu, Ruan, et~al.]{wu2024janus}
Wu~, C., Chen , X., Wu~, Z., Ma~, Y., Liu , X., Pan , Z., Liu , W., Xie , Z., Yu~, X., Ruan , C., \& others  (2024)
\newblock Janus: Decoupling visual encoding for unified multimodal understanding and generation.
\newblock \emph{arXiv preprint arXiv:2410.13848}

\bibitem[Wu et~al.(2023)Wu, Fei, Qu, Ji, and Chua]{wu2023next}
Wu~, S., Fei , H., Qu~, L., Ji~, W., \& Chua , T.-S. (2023)
\newblock Next-gpt: Any-to-any multimodal llm.
\newblock \emph{arXiv preprint arXiv:2309.05519}

\bibitem[Xiao et~al.(2024)Xiao, Zhu, Zhai, Zhou, and Zong]{xiao2024diusum}
Xiao , M., Zhu , J., Zhai , F., Zhou , Y., \& Zong , C. (2024)
\newblock Diusum: Dynamic image utilization for multimodal summarization. In
\newblock \emph{Proceedings of the AAAI Conference on Artificial Intelligence}
\newblock \emph{38}, pp. \penalty0 19297--19305.

\bibitem[Yao et~al.(2024)Yao, Yu, Zhang, Wang, Cui, Zhu, Cai, Li, Zhao, He, et~al.]{yao2024minicpm}
Yao , Y., Yu~, T., Zhang , A., Wang , C., Cui , J., Zhu , H., Cai , T., Li~, H., Zhao , W., He~, Z., \& others  (2024)
\newblock Minicpm-v: A gpt-4v level mllm on your phone.
\newblock \emph{arXiv preprint arXiv:2408.01800}

\bibitem[Ye et~al.(2024)Ye, Xu, Liu, Hu, Yan, Qian, Zhang, Huang, and Zhou]{ye2024mplug}
Ye~, J., Xu~, H., Liu , H., Hu~, A., Yan , M., Qian , Q., Zhang , J., Huang , F., \& Zhou , J. (2024)
\newblock mplug-owl3: Towards long image-sequence understanding in multi-modal large language models.
\newblock \emph{arXiv preprint arXiv:2408.04840}

\bibitem[Zauner(2010)]{zauner2010implementation}
Zauner , C. (2010)
\newblock Implementation and benchmarking of perceptual image hash functions

\bibitem[Zheng et~al.(2024)Zheng, Zhang, Zhang, Ye, Luo, Feng, and Ma]{Zheng_LlamaFactory_Unified_Efficient_2024}
Zheng , Y., Zhang , R., Zhang , J., Ye~, Y., Luo , Z., Feng , Z., \& Ma~, Y. (2024)
\newblock {LlamaFactory: Unified Efficient Fine-Tuning of 100+ Language Models}. In
\newblock Association for Computational Linguistics.

\bibitem[Zhu et~al.(2024)Zhu, Lee, Zhang, Harsha, Feujio, Maharaj, and Li]{zhu2024murar}
Zhu , Z., Lee , D., Zhang , H., Harsha , S.~S., Feujio , L., Maharaj , A., \& Li~, Y. (2024)
\newblock Murar: A simple and effective multimodal retrieval and answer refinement framework for multimodal question answering.
\newblock \emph{arXiv preprint arXiv:2408.08521}

\end{thebibliography}
\normalsize


\clearpage
\appendix
\section{Limitations\label{sec:limitations}}

Although our proposed methods outperform some existing approaches for the M$^2$RAG task, there are still several limitations.

\paragraph{More Modalities}
Our work mainly focus on the text and image modalities. 
In the real-world scenarios, information in other modalities are also crucial, such as speech and video.
In the future, we plan to extend our proposed benchmark and training datasets to more modalities, even the mixed modalities.

\paragraph{Evaluation Metrics}
The proposed evaluation metrics might fail to capture all potential aspects of M$^2$RAG evaluation. For example, pairwise ranking is more appropriate for assessing overall performance than Likert scale in M$^2$RAG task, which lacks a standard optimal answer. 
Besides, due to the requirement of using LLMs and MLLMs to evaluate the quality of long-text multi-modal content across 8 dimensions\footnote{Please refer to Table~\ref{tab:eval_costs} for more details about evaluation cost.}, this leads to huge evaluation costs in large-scale evaluation. 
To balance the affordability and reliability of experiments on our proposed benchmark, we introduce a quantity-first approach with 1,000 evaluation samples in this paper. In future work, we plan to distill and build efficient text-modal and multi-modal metrics to support larger-scale benchmark testing.

\section{Ethical Considerations\label{app:ethical}}
Most of the task inputs in our benchmark and training dataset are sourced from publicly available datasets, ensuring that they pose no harm to individuals or groups. Furthermore, the text generated by large language models (LLMs) and multi-modal language models (MLLMs) is carefully curated and processed by human annotators to safeguard privacy and confidentiality. No personally identifiable information (PII) is included. However, it is important to note that texts from the ELI5 dataset~\cite{fan2019eli5} and multi-modal documents retrieved via Google Custom Search may contain harmful content or hate speech. Despite these potential risks, it is crucial to disclose the full scope of this research, as materials from ELI5 and Google Custom Search have been extensively used in safety research within the community. All annotators are compensated fairly, with an hourly wage of approximately \$4.76 USD, which exceeds the average hourly wage of \$3.13 USD on Amazon Mechanical Turk~\cite{hara2018data}.

\section{Query Collection\label{app:query_collection}}
To simulate the real-world user query solving problem, we collect diverse and high-quality user queries from the ELI5 dataset~\cite{fan2019eli5} (The dataset is under BSD License). The ELI5 corpus is particularly suited for our task due to its comprehensive collection of long-form, open-ended questions that necessitate detailed, multi-sentence responses. The diversity of topics in ELI5 presents an opportunity to reflect the capabilities of language models in the real-world scenarios.
Subsequently, we conduct two steps to collect diverse and high-quality queries that are suitable for using multi-modal information as responses: (1) Query Filtering; and (2) Query Classification.

\begin{figure}[H]
\scriptsize
\centering
\begin{tcolorbox}
\textbf{\# Task Description}\\
You are a useful text classification assistant. You will be provided with a piece of text to be recognized. Your task is to classify the text with the guidelines.\\
\\
\textbf{\# Guidelines}\\
1. If the input text is not a question, the description is unclear, you are supposed to generate a single integer 0, which means it is an invalid question.\\
2. If there is no need to cite rich references to answer the question, i.e. the question can be fully addressed with a simple answer, you are supposed to generate a single integer 0, which means it is an invalid question.\\
3. If the question is not classified as invalid question according to the above two rules, then generate a single integer 1.\\
\\
\textbf{\# Output Format}\\
Your output should consist of two lines.\\
- The first line is your brief analysis of the input question content and why or why not it meets the requirements;\\
- The second line is an single integer 0 or 1.\\
Do not generate any other information other than the analysis and category indices.\\
\\
\textbf{\# Input Text}\\
\{text\}\\
\\
\textbf{\# Output}\\
\end{tcolorbox}
\caption{Prompt template for filtering queries which are not complex questions.}
\label{fig:query_filter_is_question}
\end{figure}

\begin{figure}[H]
\scriptsize
\centering
\begin{tcolorbox}
\textbf{\# Task Description}\\
You are a useful question classification assistant. Your task is to classify the input question with the guidelines.\\
\\
\textbf{\# Guidelines}\\
1. If inserting appropriate images in the answer to this question can provide a more comprehensive answer and make it easier for users to understand the content being discussed, then generate a single integer 1.\\
2. Otherwise, generate a single integer 0.\\
\\
\textbf{\# Output Format}\\
Your output should consist of two lines.\\
- The first line is your brief analysis of the input question content and why or why not it meets the requirements;\\
- The second line is an single integer 0 or 1.\\
Do not generate any other information other than the analysis and category indices.\\
\\
\textbf{\# Input Question}\\
\{question\}\\
\\
\textbf{\# Output}\\
\end{tcolorbox}
\caption{Prompt template for filtering queries which are not necessarily answered with images.}
\label{fig:query_filter_need_image}
\end{figure}

\paragraph{Query Filtering} conduct two steps to filter user queries that require detailed explanation and visual information for better understanding the content in the responses. 
The first step involves eliminating invalid or low-quality queries. Specifically, GPT-4o is prompted to remove queries that do not elicit detailed responses, such as those reducible to yes/no answers or simple sentences (as illustrated in Figure~\ref{fig:query_filter_is_question}). This ensures that the dataset focuses on queries requiring more elaborate, informative responses. 
The second step filters out queries that can be fully addressed through textual content alone, as illustrated in Figure~\ref{fig:query_filter_need_image}. Queries that lack the necessity for visual information do not align with the goal of evaluating multi-modal capabilities and are thus excluded.

\begin{figure}[H]
\scriptsize
\centering
\begin{tcolorbox}[sidebyside]
\textbf{\# Task Description}\\
You are a useful question classification assistant. Your task is to classify the input question with the guidelines.\\
\\
\textbf{\# Category List}\\
\textbf{1. Society \& Culture}: This category encompasses questions about traditions, languages, customs, behaviors, and societal norms. Images can serve as a visual reference to cultural landmarks, traditional attire, or historical events that support the textual explanation.\\
\textbf{Example}:  What is the significance of the Day of the Dead celebration in Mexican culture?\\
\textbf{2. Science \& Mathematics}: Questions in this category often deal with complex concepts that can be enhanced with diagrams, charts, or images. These visual aids can help in understanding scientific phenomena or mathematical theories.\\
\textbf{Example}: How does photosynthesis work in plants?\\
\textbf{3. Health}: Health-related questions often benefit from the use of images like anatomical diagrams, charts, or photos for better understanding of medical conditions, treatments, or fitness exercises.\\
\textbf{Example}: What are the symptoms and treatment options for carpal tunnel syndrome?\\
\textbf{4. Education \& Reference}: This category includes academic topics and general knowledge where explanatory images, instructional graphics, and reference tables can provide clarity and enhance learning.\\
\textbf{Example}: What are the main parts of a plant cell and their functions?\\
\textbf{5. Computers \& Internet}: Questions about technology, software, and internet usage can be more effectively answered with screenshots, infographics, and step-by-step visual guides.\\
\textbf{Example}: How do I set up a Wi-Fi network at home?\\
\textbf{6. Sports}: Sports-related queries can be explained better with visual demonstrations, play diagrams, athlete photos, and equipment images to illustrate techniques, rules, or historical moments.\\
\textbf{Example}: What are the basic rules of soccer?\\
\textbf{7. Business \& Finance}: Financial and business concepts often involve data, charts, and visualizations that can make complex information more digestible and insightful.\\
\textbf{Example}: How does compound interest work and what impact can it have on investments?\\
\textbf{8. Entertainment \& Music}: This category includes questions about movies, TV shows, music, and celebrities where images of album covers, scene snapshots, or charts of music theory can enrich the content.\\
\textbf{Example}: What was the cultural impact of the TV show `Friends'?\\

\tcblower

\textbf{9. Family \& Relationships}: Questions about family dynamics, parenting, and social relationships often benefit from illustrative images such as family diagrams, photos, and infographics about communication techniques or social scenarios.\\
\textbf{Example}: What are some effective conflict resolution strategies for couples?\\
\textbf{10. Politics \& Government}: Political and governmental questions can be clarified with the help of charts, maps, and historical photographs that provide visual context to legislative processes, electoral maps, or political events.\\
\textbf{Example}: How does the electoral college system work in the United States?\\
\textbf{11. Others}\\
\\
\textbf{\# Classification Guidelines}\\
Select one or more categories from categories 1-10 that are most relevant to the content of the question as the category of the question. If the question does not belong to any of the categories 1-10, then classify it into category 11.\\
\\
\textbf{\# Precautions}\\
1. If the input text is classified into category 11, it cannot be assigned any other category.\\
2. Please first briefly analyze the content of the input text and find its association with the given category, and then output category indices based on the above analysis.\\
\\
\textbf{\# Output Format}\\
Your output should consist of two lines.\\
- The first line is your brief analysis of the input text content and its relationship with the given category;\\
- The second line consists of a list of integers separated by a space. Each integer represents the index of a category you assigned to the input question.\\
Do not generate any other information other than the analysis and category indices.\\
\\
\textbf{\# Input Question}\\
\{question\}\\
\\
\textbf{\# Output}\\
\end{tcolorbox}
\caption{Prompt template for query classification.}
\label{fig:query_classify}
\end{figure}

\paragraph{Query Classification} To effectively analyze the difference among user query topics, we also prompt GPT-4o model to automatically categorize these queries into eleven topics from Yahoo Answers Topics\footnote{\url{https://huggingface.co/datasets/community-datasets/yahoo_answers_topics}}. Table~\ref{tab:query_categories} demonstrates that our user queries covers common and diverse searching scenarios, which is beneficial to reflecting the performance of evaluated models in real-world scenarios. The prompt template for query classification is illustrated in Figure~\ref{fig:query_classify}.

\begin{table}[H]
    \centering
    \caption{\label{tab:query_categories} Topics and volumes of the queries.}
    \resizebox{0.6\linewidth}{!}{
        \begin{tabular}{ccc}
        \toprule
            \textbf{ID} & \textbf{Category Name} & \textbf{Num. Samples} \\
        \midrule
            1 & Society \& Culture & 14182 \\
            2 & Science \& Mathematics & 14449 \\
            3 & Health & 10931 \\
            4 & Education \& Reference & 776 \\
            5 & Computers \& Internet & 6999 \\
            6 & Sports & 1974 \\
            7 & Business \& Finance & 3821 \\
            8 & Entertainment \& Music & 1946 \\
            9 & Family \& Relationships & 820 \\
            10 & Politics \& Government & 4638 \\
            11 & Others & 451 \\
        \bottomrule
        \end{tabular}
    }
\end{table}

\section{Retrieval of Collected Elements\label{app:element_relecance_eval}}
We replaced traditional embedding-based retrieval methods with LLM-based and MLLM-based approaches to improve the retrieval performance. 
Embedding-based methods typically prioritize semantic similarity but often overlook whether candidate elements provide substantial information relevant to the target question. Additionally, in real-time M$^2$RAG tasks, where online documents cannot be pre-indexed, LLM-based and MLLM-based methods offer better performance without incurring significant additional time costs.

Using these methods, text and visual elements within the collected documents are evaluated based on their relevance to the user query. This evaluation is conducted by prompting LLMs and MLLMs for textual and visual content, respectively. The prompt templates used for these evaluations are illustrated in Figures~\ref{fig:text_retrieval} and~\ref{fig:visual_retrieval}.
This step results in a score from 0 to 10 for each element, where higher scores indicate better relevancy.

\begin{figure}[H]
\scriptsize
\centering
\begin{tcolorbox}
\textbf{\# Task Description}\\
You are a useful text evaluation assistant. You will be provided with a user query and a piece of text extracted from a relevant web page. Please evaluate whether this text contains useful information to address the user query.\\
\\
\textbf{\# Input Data}\\
1. User Query\\
2. Text from a Web Page\\
\\
\textbf{\# Evaluation Guidelines}\\
Assign a score from 0 to 10, where a higher score indicates better alignment:\\
- 0: Entirely irrelevant to the query.\\
- 1-3: Minimal relevance; most content deviates from the query.\\
- 4-6: Moderately relevant; some content aligns with the query, but includes notable digressions.\\
- 7-9: Highly relevant; most content relates directly to the query with minor digressions.\\
- 10: Perfectly relevant; fully aligns with the query without any digressions.\\
\\
\textbf{\# User Query}\\
\{query\}\\
\\
\textbf{\# Web Page Text}\\
\`{}\`{}\`{}markdown\\
\{webpage\_piece\}\\
\`{}\`{}\`{}\\
\\
\textbf{\# Precautions}\\
If there is an \verb|<IMAGE_PLACEHOLDER>| in the webpage text, ignore it in your evaluation.\\
\\
\textbf{\# Output Format}\\
Your output should consist of two lines:\\
1. A brief analysis of the text's relevance to the query.\\
2. An integer score from 0 to 10.\\
\\
\textbf{\# Output}\\
\end{tcolorbox}
\caption{Prompt template for the retrieval of text elements.}
\label{fig:text_retrieval}
\end{figure}

\begin{figure}[H]
\scriptsize
\centering
\begin{tcolorbox}
\textbf{\# Task Description}\\
You are an assistant for evaluating the correlation between image and text. You will be provided with an image and a question. The image comes from a web page, so it may be a highly relevant image to the provided text, a slightly relevant image, or a completely unrelated image, or even an advertising image.\\
\\
\textbf{\# Scoring Strategy}\\
Score from 0 to 10 based on the correlation between the image content and the problem, with higher scores indicating higher correlation.\\
- 0 represents that the image is completely unrelated to the question, or the image is an advertising image.\\
- 1-3 represents that the image is slightly related to the question, and is dispensable for answering the question.\\
- 4-6 represents that the image is moderately related to the question, and it is helpful for answering some of the content in the question.\\
- 7-9 represents that the image is highly related to the question, and it is very helpful for answering the question.\\
- 10 represents that the image is the key to the question, and it is essential for answering the question.\\
\\
\textbf{\# Precautions}\\
You are not expected to generate any other information other than the score (an integer).\\
\\
\textbf{\# Question}\\
\{question\}\\
\\
\textbf{\# Score}\\
\end{tcolorbox}
\caption{Prompt template for the retrieval of visual elements.}
\label{fig:visual_retrieval}
\end{figure}

\section{Dataset Statistics\label{app:dataset_statistics}}

\subsection{Basic Statistics}

The statistical information about our benchmark are shown in Table~\ref{tab:score_statistics} and~\ref{tab:data_statistics}.

\begin{table}[H]
    \centering
    \caption{The average values and standard deviations for element scores of the full benchmark dataset.}
    \label{tab:score_statistics}
    \resizebox{0.4\linewidth}{!}{
        \begin{tabular}{ccc}
            \toprule
                \textbf{Item} & \textbf{Avg. Score} & \textbf{Std.} \\
            \midrule
                \textbf{Text} & 1.5 & 2.3 \\
                \textbf{Image} & 3.2 & 2.3 \\
                \textbf{Aux. Image} & 3.8 & 2.6 \\
            \bottomrule
        \end{tabular}
    }
\end{table}

\begin{table}[H]
    \centering
    \caption{Numerical statistics of the full benchmark dataset.}
    \label{tab:data_statistics}
    \resizebox{0.6\linewidth}{!}{
        \begin{tabular}{cccc}
            \toprule
                \textbf{Item} & \textbf{Range} & \textbf{Avg. Num} & \textbf{Std.} \\
            \midrule
                \textbf{Web Page} & \multirow{3}{*}{per query} & 9.8 & 1.0 \\
                \textbf{Image} & & 12.0 & 16.7 \\
                \textbf{Aux. Image} & & 3.0 & 2.8 \\
            \midrule
                \textbf{Text Element} & \multirow{2}{*}{per web page} & 25.7 & 43.2 \\
                \textbf{Image} & & 1.2 & 4.9 \\
            \bottomrule
        \end{tabular}
    }
\end{table}

\subsection{Data Distribution}
\begin{figure}[H]
    \centering
    \begin{minipage}{0.8\linewidth}
        \centering
        \includegraphics[width=0.5\linewidth]{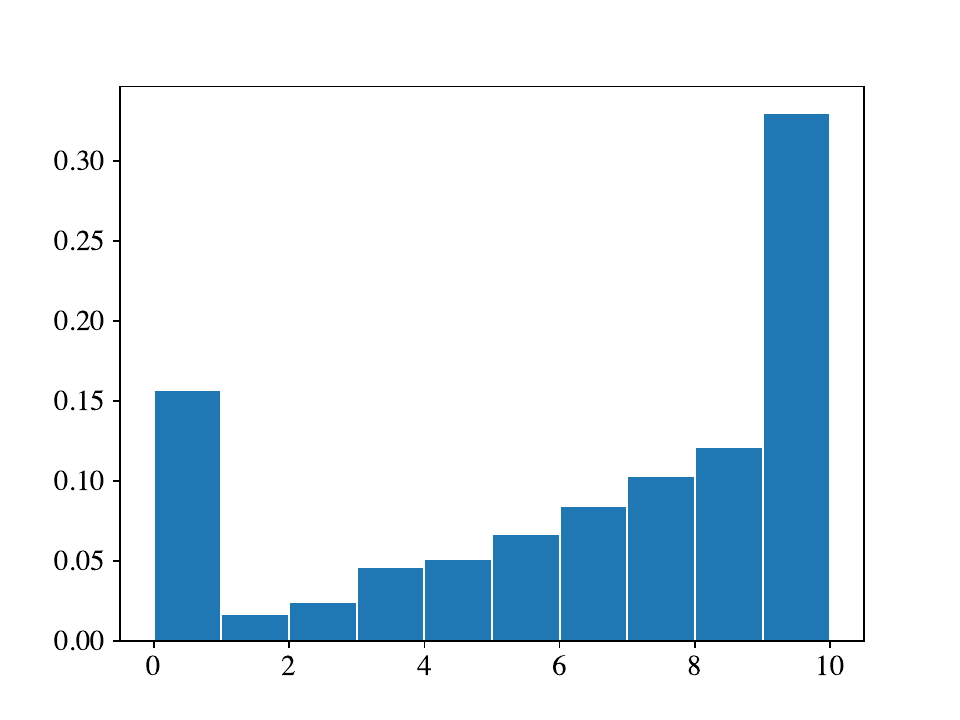}
        \caption{Non-empty web pages for each query}
        \label{fig:num_webpage}
    \end{minipage}
    
    \begin{minipage}{0.4\linewidth}
        \centering
        \includegraphics[width=1.0\linewidth]{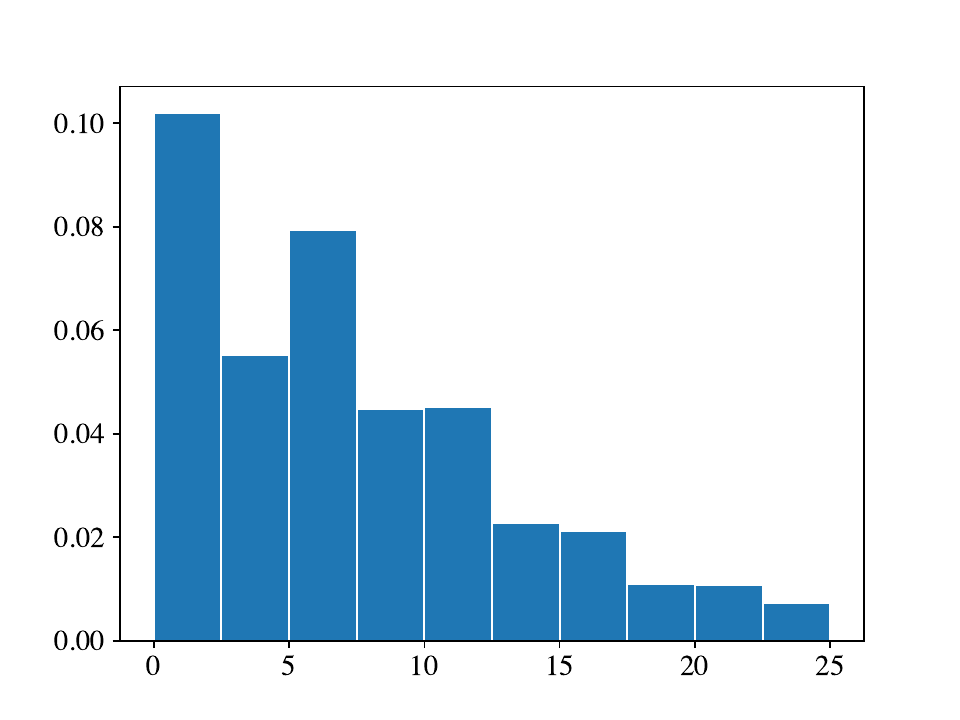}
        \caption{Valid pieces in each web page}
        \label{fig:dist_valid_piece}
    \end{minipage}
    \begin{minipage}{0.4\linewidth}
        \centering
        \includegraphics[width=1.0\linewidth]{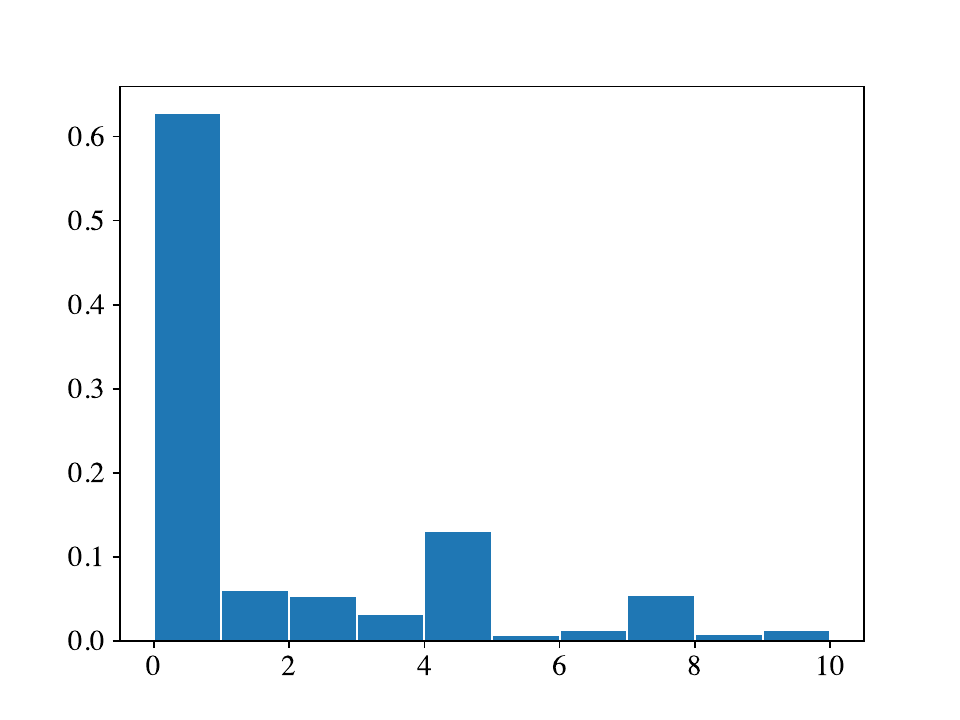}
        \caption{Scores of web page pieces}
        \label{fig:dist_piece_score}
    \end{minipage}

    \begin{minipage}{0.4\linewidth}
        \centering
        \includegraphics[width=1.0\linewidth]{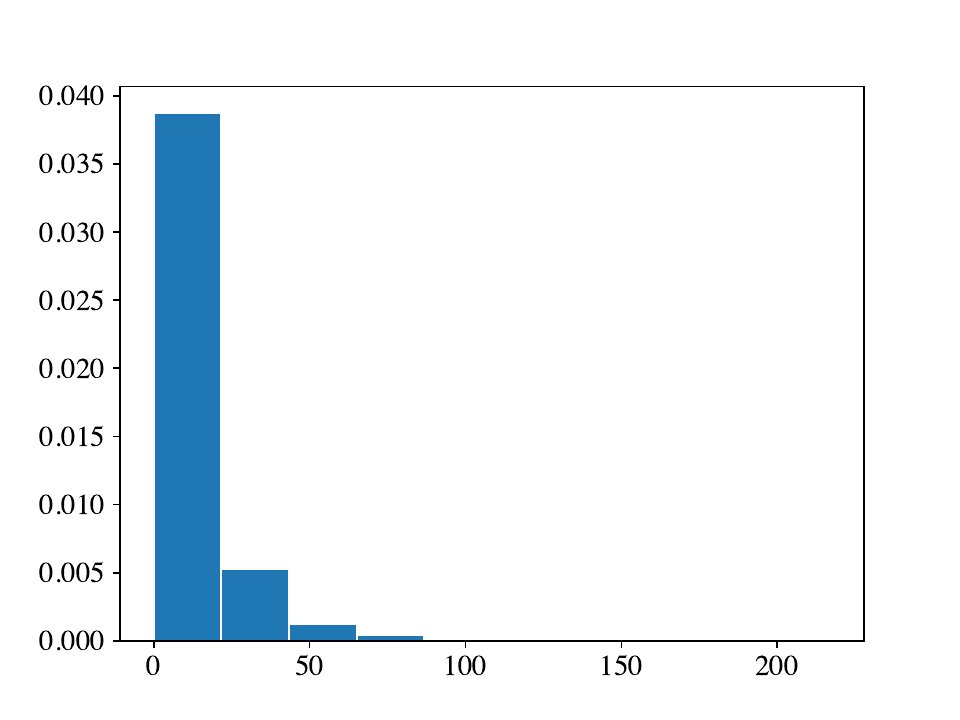}
        \caption{Web page images for each query}
        \label{fig:dist_web_page_image}
    \end{minipage}
    \begin{minipage}{0.4\linewidth}
        \centering
        \includegraphics[width=1.0\linewidth]{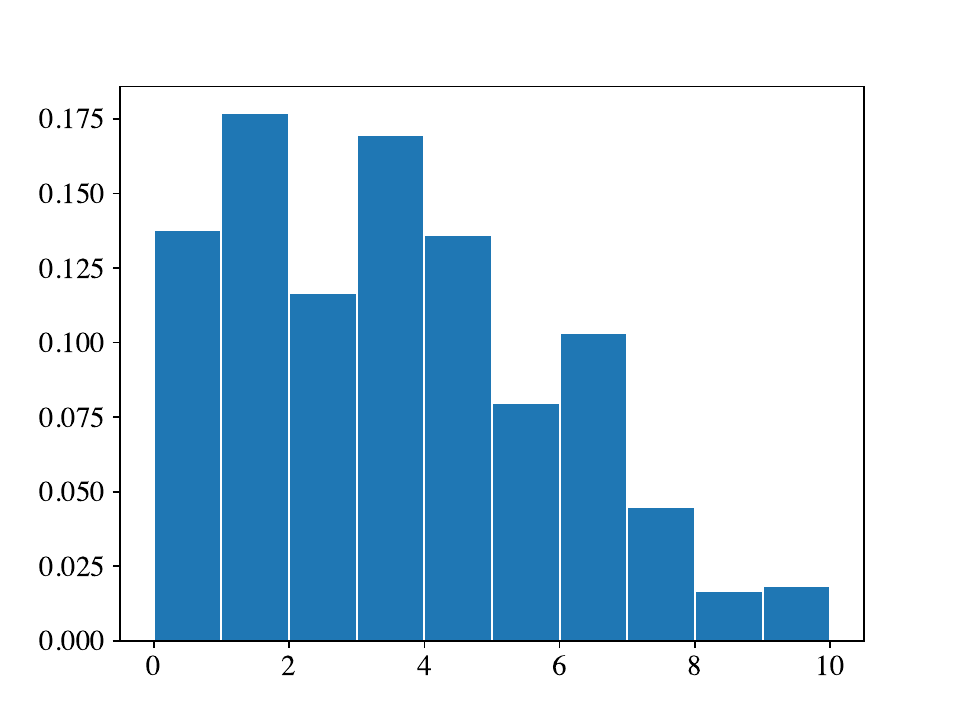}
        \caption{Score of web page images}
        \label{fig:dist_web_page_image_score}
    \end{minipage}
    
    \begin{minipage}{0.4\linewidth}
        \centering
        \includegraphics[width=1.0\linewidth]{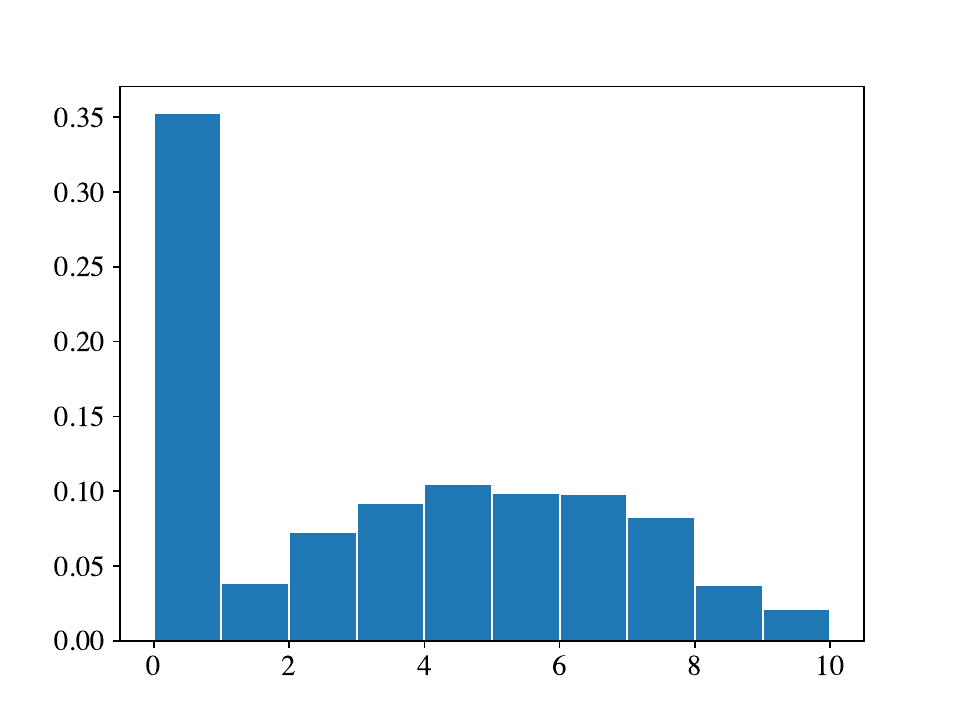}
        \caption{Auxiliary images for each query}
        \label{fig:dist_aux_image}
    \end{minipage}
    \begin{minipage}{0.4\linewidth}
        \centering
        \includegraphics[width=1.0\linewidth]{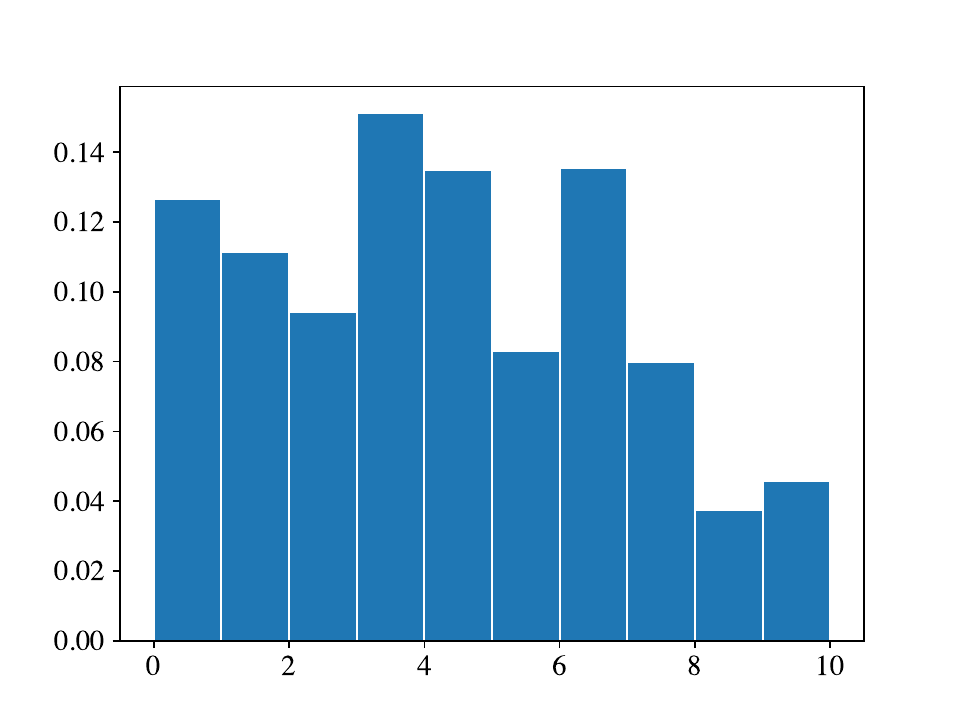}
        \caption{Score of auxiliary images}
        \label{fig:dist_aux_image_score}
    \end{minipage}
\end{figure}

We list the distributions of some key items:
\begin{itemize}
    \item The number of non-empty web pages for each query, as illustrated in Figure~\ref{fig:num_webpage};
    \item The number of valid text pieces in each web page, as illustrated in Figure~\ref{fig:dist_valid_piece};
    \item The score of each text piece, as illustrated in Figure~\ref{fig:dist_piece_score};
    \item The number of images in web pages for each query, as illustrated in Figure~\ref{fig:dist_web_page_image};
    \item The score of each web page image, as illustrated in Figure~\ref{fig:dist_web_page_image_score};
    \item The number of auxiliary images for each query, as illustrated in Figure~\ref{fig:dist_aux_image};
    \item The score of each auxiliary image, as illustrated in Figure~\ref{fig:dist_aux_image_score}.
\end{itemize}

\section{Details of Evaluation Metrics\label{app:eval_metrics}}

\subsection{Prompt Templates}
Apart from Context Precision and Faithfulness which are implemented with RAGAS, other metrics are implemented by prompting GPT-4o with the following prompt templates:
\begin{itemize}
    \item Fluency: as illustrated in Figure~\ref{fig:eval_text_fluency};
    \item Relevance: as illustrated in Figure~\ref{fig:eval_text_relevance};
    \item Image Coherence: as illustrated in Figure~\ref{fig:eval_image_coherence};
    \item Image Helpfulness: as illustrated in Figure~\ref{fig:eval_image_helpfulness};
    \item Image Reference: as illustrated in Figure~\ref{fig:eval_image_reference};
\end{itemize}

\begin{figure}[H]
\scriptsize
\centering
\begin{tcolorbox}
\textbf{\# Task Description}\\
You are a document evaluation assistant. Your task is to evaluate the fluency of the provided text written in markdown.\\
\\
\textbf{\# Input Text}\\
\`{}\`{}\`{}markdown\\
\{text\}\\
\`{}\`{}\`{}\\
\\
\textbf{\# Scoring Criteria}\\
Assign a score from 0 to 10 based on the fluency of the text:\\
- \textbf{0:} The text is incoherent, lacks logical flow, and deviates significantly from correct markdown syntax.\\
- \textbf{1-3:} The text has potential with partial fluency and logic but is plagued by multiple markdown errors, such as unhandled duplicate headings or incorrect formatting structures.\\
- \textbf{4-6:} The text demonstrates general fluency with minor grammatical or logical inconsistencies, though the markdown structure lacks clarity or uniformity, like having redundant sections.\\
- \textbf{7-9:} The text is well-written and logically coherent, with proper markdown usage, forming a mostly seamless document that maintains consistent quality.\\
- \textbf{10:} The text is exemplary, demonstrating perfect fluency, logical progression, and impeccable markdown syntax, representing an ideal markdown document.\\
\\
Be rigorous and discerning when assigning your score.\\
\\
\textbf{\# Output Instructions}\\
Your output should contain only 2 lines:\\
1. A brief explanation justifying the assigned score.\\
2. The score as an integer value.\\
\\
Do not provide any additional information beyond the explanation and score.\\
\end{tcolorbox}
\caption{Prompt template for the text fluency evaluation.}
\label{fig:eval_text_fluency}
\end{figure}

\begin{figure}[H]
\scriptsize
\centering
\begin{tcolorbox}
\textbf{\# Task Description}\\
You are a document evaluation assistant. Your task is to evaluate the relevance between the query and provided text written in markdown.\\
\\
\textbf{\# Query}\\
\{query\}\\
\\
\textbf{\# Input Text}\\
\`{}\`{}\`{}markdown\\
\{text\}\\
\`{}\`{}\`{}\\
\\
\textbf{\# Scoring Criteria}\\
Assign a score from 0 to 10 based on the degree of relevance between the input text and the given query:\\
- \textbf{0:} The text is entirely irrelevant to the given query.\\
- \textbf{1-3:} The text has minimal relevance, with most content deviating from the intended query.\\
- \textbf{4-6:} The text is moderately relevant, with some content aligned with the intended query, but contains noticeable digressions.\\
- \textbf{7-9:} The text is highly relevant, with most content directly related to the intended query and only minor digressions.\\
- \textbf{10:} The text is perfectly relevant, fully aligning with the intended query without any digressions.\\
\\
Be rigorous and discerning when assigning your score.\\
\\
\textbf{\# Output Instructions}\\
Your output should contain only 2 lines:\\
1. A brief explanation justifying the assigned score.\\
2. The score as an integer value.\\
\\
Do not provide any additional information beyond the explanation and score.\\
\end{tcolorbox}
\caption{Prompt template for the text relevance evaluation.}
\label{fig:eval_text_relevance}
\end{figure}

\begin{figure}[H]
\scriptsize
\centering
\begin{tcolorbox}[sidebyside]
\textbf{\# Task Description}\\
You are a multimodal document evaluation assistant. You will receive an image and its textual context written in markdown. Your task is to evaluate the coherence between the image and the text (context above and below) it accompanies.\\
\\
\textbf{\# Context Above}\\
\`{}\`{}\`{}markdown\\
\{context\_above\}\\
\`{}\`{}\`{}\\
\\
\textbf{\# Image}\\
\verb|<IMAGE_PLACEHOLDER>|\\
\\
\textbf{\# Context Below}\\
\`{}\`{}\`{}markdown\\
\{context\_below\}\\
\`{}\`{}\`{}\\
\\
\textbf{\# Scoring Criteria and Examples}\\
Evaluate how well the image complements its accompanying text and assign a score from 0 to 10. A higher score reflects stronger coherence between the image and the text. Be precise in your evaluation.\\
\\
- \textbf{Score 0-3}: Minimal or no coherence between the image and text.\\
\text{\quad}- \textit{Example 1}: The image shows a video game controller, but the text discusses supply and demand in virtual goods on platforms like Steam. These topics are unrelated, and the image adds no context to the discussion of virtual goods. The score should be 0-3.\\
\text{\quad}- \textit{Example 2}: The image depicts a luxury car, but the text is about the environmental impact of electric vehicles. While both relate to automobiles, the image doesn’t contribute to the discussion of electric vehicles’ environmental effects. Thus, it merits a score of 0-3.\\
\text{\quad}- \textit{Example 3}: A photo of a pizza is shown, but the text is a deep dive into the history of ancient Greece. There is no connection between the image and the historical content, making the coherence score 0-3.\\
\\
- \textbf{Score 4-6}: Some coherence, but with unrelated elements.\\
\text{\quad}- \textit{Example 1}: The image shows various components like a remote control, cables, and a media player, which are related to the media player mentioned in the text. However, the context focuses on video formats, which is not addressed by the image. The coherence score should be between 4 and 6.\\
\text{\quad}- \textit{Example 2}: The image shows graduates from Harvard University, and the text is about the cost of university education and the perceived benefits of attending prestigious institutions. While the image relates to the general topic of education, it doesn’t directly support the discussion about the financial costs or value of education, meriting a score of 4-6.\\

\tcblower

\text{\quad}- \textit{Example 3}: The image of a crowded subway is referenced in the text discussing urbanization and its effects on public transport. However, the text focuses more on the environmental impact of urbanization rather than directly linking to the image of subway crowds, leading to a coherence score of 4-6.\\
- \textbf{Score 7-9}: High coherence, with the image closely aligning with the text.\\
\text{\quad}- \textit{Example 1}: The image is a photo of a video game controller, and the text describes features of the latest gaming console. The image directly aligns with the content, visually representing the technology described in the text, earning a score of 7-9.\\
\text{\quad}- \textit{Example 2}: The image depicts a close-up of a laptop keyboard, and the text discusses advancements in laptop design. The image complements the text well, helping the reader understand the specific design features mentioned, justifying a score of 7-9.\\
\text{\quad}- \textit{Example 3}: A chart showing sales data for various smartphone models is paired with text explaining smartphone market trends. The image is directly relevant to the discussion of market share and trends, aiding comprehension, and warrants a score of 7-9.\\
\\
- \textbf{Score 10}: Perfect coherence, where the image completely corresponds to and enhances the text.\\
\text{\quad}- \textit{Example 1}: The image is a detailed picture of a video game controller, and the text provides a comprehensive review of the controller’s features, design, and performance. The image fully complements the text, making the reader’s understanding more complete and vivid, deserving a score of 10.\\
\text{\quad}- \textit{Example 2}: The image shows a close-up of a public toilet seat, which is relevant to the context below that discusses the design differences between public and home toilet seats. The image is completely aligned with the text, directly aiding the discussion. The score should be 10.\\
\text{\quad}- \textit{Example 3}: The image is an anatomical diagram of the human heart, and the text explains how blood circulates through the body. The visual directly supports the explanation, making the text easier to understand and reinforcing key points, justifying a score of 10.\\
\\
\textbf{Note}: The image needs to be coherent with the context above OR below it, not necessarily both. Be rigorous and objective in your assessment.\\
\\
\textbf{\# Output Instructions}\\
Your output should contain only 2 lines:\\
1. A brief explanation justifying the assigned score.\\
2. The score as an integer value.\\
\\
Do not provide any additional information beyond the explanation and score.\\
\end{tcolorbox}
\caption{Prompt template for the image coherence evaluation.}
\label{fig:eval_image_coherence}
\end{figure}

\begin{figure}[H]
\scriptsize
\centering
\begin{tcolorbox}[sidebyside]
\textbf{\# Task Description}\\
You are a multimodal document evaluation assistant. You will receive an image and its textual context written in markdown. Your task is to evaluate the helpfulness of the image in enabling human readers to comprehend the text (context above and below) it accompanies.\\
\\
\textbf{\# Context Above}\\
\`{}\`{}\`{}markdown\\
\{context\_above\}\\
\`{}\`{}\`{}\\
\\
\textbf{\# Image}\\
\verb|<IMAGE_PLACEHOLDER>|\\
\\
\textbf{\# Context Below}\\
\`{}\`{}\`{}markdown\\
\{context\_below\}\\
\`{}\`{}\`{}\\
\\
\textbf{\# Scoring Criteria and Examples}\\
Assess the image's helpfulness in improving comprehension of the text, assigning a score from 0 to 10. A higher score indicates the image significantly enhances understanding. Be precise and rigorous when assigning the score.\\
\\
- \textbf{Score 0-3}: The image is minimally or not at all helpful for comprehension.\\
\text{\quad}- \textit{Example 1}: The image depicts a book cover titled "When God Asks You to Do Something You Don't Want to Do," with a background of crashing ocean waves. The text is about jury members addressing trial observations. The image is irrelevant and unhelpful, meriting a score of 0-3.\\
\text{\quad}- \textit{Example 2}: The image shows a close-up of banknotes (euro), arranged in a fan-like pattern. While the image is related to the concept of money, it does not directly enhance the comprehension of the text, which discusses the exchange rate of different currencies. Thus, it is not helpful for comprehension, earning a score of 0-3.\\
\text{\quad}- \textit{Example 3}: The image is of a person looking at a blank computer screen, with no connection to the text discussing how the internet has influenced modern literature. The image does not provide any useful context or additional understanding of the text, so it warrants a score of 0-3.\\
\\
- \textbf{Score 4-6}: The image provides some helpful context but may include extraneous or less relevant details.\\
\text{\quad}- \textit{Example 1}: The image shows a man at a desk, shocked at his laptop, with colorful text and symbols in the background. The text discusses how hackers are caught. While the image is thematically related, it is not directly helpful in explaining the topic of cyber security, earning a score of 4-6.\\
\text{\quad}- \textit{Example 2}: The image shows a car with a vented hood, which relates to the text discussing the aerodynamic and cooling benefits of such designs in mid-engine cars. However, the image doesn’t address specific details, like material considerations or the pressure zones discussed in the text, meaning it provides partial context but not a complete understanding. Therefore, it should be assigned a score of 4-6.\\

\tcblower

\text{\quad}- \textit{Example 3}: The image of a group of college students at a campus event is referenced in the text discussing the benefits of a liberal arts education. While it offers a visual representation of student life, it doesn’t fully enhance the understanding of the specific academic benefits described, so it earns a score of 4-6.\\
\\
- \textbf{Score 7-9}: The image is highly helpful in enhancing comprehension of the text.\\
\text{\quad}- \textit{Example 1}: The image features a table comparing popular browsers and their operating systems, such as Edge, Safari, and Chrome. The text discusses why Internet Explorer (IE) is disliked. The image is highly relevant, providing a clear visual context that directly supports the reader’s understanding of the browser landscape, justifying a score of 7-9.\\
\text{\quad}- \textit{Example 2}: The image is a close-up of a laptop keyboard, with highlighted keys. The text discusses the evolution of laptop designs, and the image directly supports the explanation, helping the reader understand the physical changes over time, which justifies a score of 7-9.\\
\text{\quad}- \textit{Example 3}: The image shows the front page of a scientific journal, with headlines and an image illustrating the main topic. The text explains the content of the journal in detail, and the image directly helps readers better visualize the article’s focus, justifying a score of 7-9.\\
\\
- \textbf{Score 10}: The image perfectly enhances and clarifies the text.\\
\text{\quad}- \textit{Example 1}: The image is a detailed diagram of the human digestive system, showing organs like the stomach and intestines. The text explains the process of digestion, and the diagram complements this explanation perfectly, enhancing the reader's understanding of the content. The image and text work together seamlessly, earning a score of 10.\\
\text{\quad}- \textit{Example 2}: The image shows a close-up of a public toilet seat, which is relevant to the context below that discusses the design differences between public and home toilet seats. The image directly complements the text, providing the exact visual context needed to understand the design differences. The score should be 10.\\
\text{\quad}- \textit{Example 3}: The image is a map highlighting the trade routes discussed in the text about ancient civilizations. The map directly enhances the comprehension of the trade dynamics and locations mentioned in the text, offering a visual aid that clarifies the complex subject matter, thus earning a score of 10.\\
\\
\textbf{Note}: The image needs to be helpful for the context above OR below it, not necessarily both. Be rigorous and objective in your assessment.\\
\\
\textbf{\# Output Instructions}\\
Your output should contain only 2 lines:\\
1. A brief explanation justifying the assigned score.\\
2. The score as an integer value.\\
\\
Do not provide any additional information beyond the explanation and score.\\
\end{tcolorbox}
\caption{Prompt template for the image helpfulness evaluation.}
\label{fig:eval_image_helpfulness}
\end{figure}

\begin{figure}[H]
\scriptsize
\centering
\begin{tcolorbox}[sidebyside]
\textbf{\# Task Description}\\
You are a multimodal document quality assessment assistant. You will receive an image and its accompanying textual context, formatted in markdown. Your task is to determine whether the image is explicitly referenced or explained within the surrounding text (both above and below the image).\\
\\
\textbf{\# Context Above}\\
\`{}\`{}\`{}markdown\\
\{context\_above\}\\
\`{}\`{}\`{}\\
\\
\textbf{\# Image}\\
\verb|<IMAGE_PLACEHOLDER>|\\
\\
\textbf{\# Context Below}\\
\`{}\`{}\`{}markdown\\
\{context\_below\}\\
\`{}\`{}\`{}\\
\\
\textbf{\# Scoring Criteria}\\
Determine how well the image is referenced or explained in the surrounding text, assigning a score from 0 to 10:\\
\\
- \textbf{Score 0}: The image is not mentioned or referenced in the text.\\
\text{\quad}- \textit{Example 1}: The image is present, but there is no mention or acknowledgment of it in either the context above or below.\\
\text{\quad}- \textit{Example 2}: The image shows a scenic mountain view, but the text is entirely about office workflows, with no reference to or explanation of the image.\\
\text{\quad}- \textit{Example 3}: The image is of a fruit basket, but the text is about urban planning, without any reference to the image or its context.\\
\\
- \textbf{Score 1-3}: The image is referenced implicitly, but the reference is inapparent, improper, or incorrect.\\
\text{\quad}- \textit{Example 1}: The text discusses the Millennium Falcon’s hyperdrive system and kyber crystal energy field, which are depicted in the image. However, the image is not explicitly mentioned, and the connection is weak. The score is 1-3.\\
\text{\quad}- \textit{Example 2}: The text mentions "the latest in gaming technology," and the image shows a futuristic gaming console. The image is not directly referenced, and the connection to the text is not clearly established. The score is 1-3.\\
\text{\quad}- \textit{Example 3}: The text talks about space exploration and mentions rockets, but the image is a close-up of a satellite. The connection between the satellite and the discussion of space exploration is implied but not directly referenced. Thus, the score is 1-3.\\
\\
- \textbf{Score 4-6}: The image is referenced implicitly or explicitly, but the reference is improper or partially relevant.\\

\tcblower

\text{\quad}- \textit{Example 1}: The image shows a skier, and the text explicitly references it, but the discussion focuses on ski races, which is only loosely related to the skier's role. The reference is valid but not fully accurate, so the score is 4-6.\\
\text{\quad}- \textit{Example 2}: The image shows a tree with roots exposed, and the text references “the importance of deep roots” in discussing personal growth. While the image supports the metaphor, the connection is not fully fleshed out, leading to a score of 4-6.\\
\text{\quad}- \textit{Example 3}: The image is of a sunset, and the text discusses the environmental impact of light pollution. While the sunset relates to the topic of light, it does not directly contribute to the explanation of light pollution, resulting in a score of 4-6.\\
\\
- \textbf{Score 7-9}: The image is explicitly referenced in a generally proper and correct manner.\\
\text{\quad}- \textit{Example 1}: The image depicts a box of chips and is explicitly mentioned in the text at an appropriate point. While the reference is accurate, it might feel a little stiff, leading to a score of 7-9.\\
\text{\quad}- \textit{Example 2}: The text discusses the structural components of a building, and the image shows a detailed diagram of the beams and supports. The reference is clear and mostly accurate, with a slight stiffness in phrasing, justifying a score of 7-9.\\
\text{\quad}- \textit{Example 3}: The image depicts a city map, and the text discusses the layout of different districts. The map is explicitly referenced and properly aids the understanding of the text, but the reference could be smoother. The score should be 7-9.\\
\\
- \textbf{Score 10}: The image is explicitly referenced with complete accuracy and proper placement.\\
\text{\quad}- \textit{Example 1}: The image is explicitly mentioned and thoroughly explained in the text, with a discussion of NFL play calls perfectly aligned with the visual content. The placement and explanation of the image are spot-on, so it warrants a score of 10.\\
\text{\quad}- \textit{Example 2}: The image shows a detailed diagram of the human circulatory system, and the text provides an in-depth explanation of how blood flows through the heart and veins. The image and text work in perfect harmony to enhance the reader’s understanding. The score should be 10.\\
\text{\quad}- \textit{Example 3}: The image depicts a vintage car and is referenced explicitly in the context, which discusses the history and evolution of automobile designs. The reference is fully accurate, and the image provides essential clarity to the reader, justifying a perfect score of 10.\\
\\
\textbf{Note}: The image only needs to be referenced or explained by its context above OR below, not necessarily both. Be rigorous and objective in your assessment.\\
\\
\textbf{\# Output Instructions}\\
Your output should consist of only two lines:\\
1. A brief explanation justifying the assigned score.\\
2. The score as an integer value.\\
\\
Refrain from providing any additional information beyond the explanation and score.\\
\end{tcolorbox}
\caption{Prompt template for the image reference evaluation.}
\label{fig:eval_image_reference}
\end{figure}

\subsection{Evaluation Costs\label{app:eval_costs}}
The average costs of our evaluation metrics are listed in Table~\ref{tab:eval_costs}.

\begin{table}[H]
    \centering
    \caption{Evaluation costs of our metrics. The prices represent the average evaluation cost of the experiments using the joint modeling approach, as presented in Table ~\ref{tab:evaluation_results}. The text-modal metrics correspond to the evaluation cost of an entire generated response, whereas the multi-modal metrics correspond to the evaluation cost per image in the output. The \textbf{Context Precision} and \textbf{Faithfulness} metrics are based on the commercial model GPT-4o-mini, while all other metrics use GPT-4o.}
    \label{tab:eval_costs}
    \resizebox{0.4\linewidth}{!}{
        \begin{tabular}{cc}
            \toprule
                \textbf{Metric} & \textbf{Cost (\$ per 1K samples)} \\
            \midrule
                \rowcolor{gray!20} \multicolumn{2}{l}{\textbf{Text-modal Metrics}} \\
            \midrule
                \textbf{Fluency} & 3.50 \\
                \textbf{Relevance} & 3.58 \\
                \textbf{Context Precision} & 4.61 \\
                \textbf{Faithfulness} & 3.98 \\
            \midrule
                \rowcolor{gray!20} \multicolumn{2}{l}{\textbf{Multi-modal Metrics}} \\
            \midrule
                \textbf{Image Coherence} & 4.76 \\
                \textbf{Image Helpfulness} & 4.92 \\
                \textbf{Image Reference} & 4.70 \\
            \bottomrule
        \end{tabular}
    }
\end{table}

\section{Experimental Setup\label{app:experimental_setup}}
\paragraph{Evaluated Models}
We evaluate the capabilities of some advanced open-source and closed-source LLMs and MLLMs for M$^2$RAG task. The involved models are listed in Table \ref{tab:baseline_models}.
Besides, we also implement a separate approach, \textit{i.e.,} the MuRAR method with GPT-4o as the backbone model~\cite{zhu2024murar}.

\begin{table}[H]
    \centering
    \caption{\label{tab:baseline_models}LLMs and MLLMs used for M$^2$RAG task.}
    \resizebox{0.6\linewidth}{!}{
        \begin{tabular}{cc}
        \toprule
            \textbf{LLM} & \textbf{MLLM} \\
        \midrule
            GPT-4o & GPT-4o \\
            DeepSeek-V3 & Step-1o \\
            Llama-3.1-70B-Instruct & Llama-3.2-90B-Vision-Instruct \\
            Llama-3.1-8B-Instruct & Llama-3.2-11B-Vision-Instruct \\
            Qwen-2.5-72B-Instruct & Qwen-2-VL-72B-Instruct \\
            Qwen-2.5-7B-Instruct & Qwen-2-VL-7B-Instruct \\
        \bottomrule
        \end{tabular}
    }
\end{table}

\paragraph{Implementation Details}
For the generation process, we deployed open-source LLMs and MLLMs using vLLM~\cite{kwon2023efficient} on Nvidia A100-SXM4-80GB GPUs. Models with 7B-11B parameters operate on a single GPU, while those exceeding 70B parameters are distributed across 4 GPUs using tensor parallelism. The context length is configured at 64k for the Llama series and 32k for the Qwen series.
During generation, we use a top-$k$ selection method with $k$ set to 20 for filtering text elements from each webpage. The maximum number of auxiliary and web page specific images is restricted to 5, with a total input image cap at 10. Images are resized to 512 $\times$ 512 thumbnails for MLLMs processing.
For evaluation, we implemented our customized metrics using the GPT-4o model to ensure the robust comprehensive assessment. 
Besides, the RAG metrics are implemented with RAGAS by using the GPT-4o-mini model.

\section{Implementation Details of Model Distillation\label{app:sft_settings}}
We fine-tune the open-source LLMs Qwen2.5-7B-Instruct, Llama-3.1-8B-Instruct and MLLM Qwen2-VL-7B-Instruct to serve as the distilled models. To ensure the fine-tuned model effectively captures the comprehensive information embedded in the training corpus, we set the context length to 32,768 tokens during fine-tuning, accommodating the majority of samples within the dataset. To optimize the computational efficiency and uphold the performance of the fine-tuned model, we employed Low-Rank Adaptation (LoRA)~\cite{hu2021lora} with the rank of 128 and $\alpha$ of 256. Apart from that, we adopt various methods to accelerate training including ZeRO~\cite{rajbhandari2020zero}, Flash Attention 2~\cite{dao2023flashattention2} and Liger Kernel~\cite{hsu2024ligerkernelefficienttriton}. The model training was conducted on 2 Nvidia A100-SXM4-80GB GPUs with a global batch size of 64 over 3 epochs, resulting in a total of 75 training steps. All models are fine-tuned with LLaMA-Factory framework~\cite{Zheng_LlamaFactory_Unified_Efficient_2024}.

\section{Results of Representative Models on the Entire Benchmark Dataset\label{app:all_results}}

The evaluation results of several representative models on the entire benchmark dataset are listed in Table~\ref{tab:eval_results_entire_dataset}.

\begin{table}[H]
    \tiny
    \centering
    \caption{The overall experiment results of representative models on the entire benchmark dataset.}
    \label{tab:eval_results_entire_dataset}
    \resizebox{1.0\linewidth}{!}{
        \begin{tabular*}{0.955\linewidth}{cclccccccccc}
        \toprule
            \multirow{2}{*}[-1.0ex]{\textbf{\begin{tabular}[c]{@{}c@{}}Model\\Type\end{tabular}}} & \multirow{2}{*}[-1.0ex]{\textbf{\begin{tabular}[c]{@{}c@{}}Generation\\Strategy\end{tabular}}} & \multirow{2}{*}[-1.0ex]{\textbf{Model}} & \multicolumn{4}{c}{\textbf{Text-modal Metrics}} & \multicolumn{4}{c}{\textbf{Multi-modal Metrics}} & \multirow{2}{*}[-1.0ex]{\textbf{Overall}} \\
        \cmidrule{4-11}
            & & & \textbf{Flu.} & \textbf{Rel.} & \textbf{CP.} & \textbf{Faith.} & \textbf{Coher.} & \textbf{Help.} & \textbf{Ref.} & \textbf{Recall} & \\
        \midrule
            \multirow{4}{*}[-1.0ex]{\textbf{Reasoners}} & \multirow{2}{*}{\textbf{Single}} & OpenAI o3-mini & 83.8 & \textbf{90.7} & 86.7 & \textbf{86.5} & 76.1 & \textbf{65.6} & \textbf{70.2} & 85.2 & \textbf{80.6} \\
            & & DeepSeek-R1 & \textbf{86.5} & 87.7 & \textbf{87.1} & 77.8 & \textcolor{red}{\textbf{78.2}} & 62.9 & 53.3 & \textbf{88.8} & 77.8 \\
        \cmidrule{2-12}
            & \multirow{2}{*}{\textbf{Multi}} & OpenAI o3-mini & 82.6 & \textbf{85.6} & \textbf{90.9} & \textbf{81.6} & 72.4 & 63.2 & 81.4 & \textcolor{red}{\textbf{100.0}} & \textbf{82.2} \\
            & & DeepSeek-R1 & \textbf{83.0} & 85.4 & 79.5 & 71.9 & \textbf{77.6} & \textcolor{red}{\textbf{69.9}} & \textbf{81.8} & 99.9 & 81.1 \\
        \midrule
            \multirow{4}{*}[-1.0ex]{\textbf{LLMs}} & \multirow{2}{*}{\textbf{Single}} & GPT-4o & \textbf{85.1} & 88.8 & \textbf{90.1} & \textcolor{red}{\textbf{86.9}} & \textbf{72.0} & \textbf{60.9} & \textbf{47.2} & 86.4 & \textbf{77.2} \\
            & & DeepSeek-V3 & 84.9 & \textbf{89.6} & 85.6 & 84.4 & 67.8 & 55.6 & 44.4 & \textbf{89.8} & 75.3 \\
        \cmidrule{2-12}
            & \multirow{2}{*}{\textbf{Multi}} & GPT-4o & \textbf{83.8} & 86.6 & \textbf{91.4} & 83.4 & \textbf{72.0} & \textbf{63.0} & \textbf{80.8} & \textcolor{red}{\textbf{100.0}} & \textbf{82.6} \\
            & & DeepSeek-V3 & 83.7 & \textbf{86.9} & 90.0 & \textbf{85.2} & 70.7 & 61.8 & 79.7 & 99.9 & 82.2 \\
        \midrule
            \multirow{4}{*}[-1.0ex]{\textbf{MLLMs}} & \multirow{2}{*}{\textbf{Single}} & GPT-4o & 85.1 & 89.1 & \textbf{89.1} & 85.9 & \textbf{71.6} & \textbf{60.2} & 46.2 & 87.4 & 76.8 \\
            & & Step-1o & \textcolor{red}{\textbf{88.2}} & \textcolor{red}{\textbf{91.1}} & \textbf{89.1} & \textbf{86.5} & 68.8 & 56.4 & \textbf{48.7} & \textbf{88.9} & \textbf{77.2} \\
        \cmidrule{2-12}
            & \multirow{2}{*}{\textbf{Multi}} & GPT-4o & 83.5 & 87.5 & 91.1 & 81.9 & 70.9 & 61.6 & 80.4 & \textcolor{red}{\textbf{100.0}} & 82.1 \\
            & & Step-1o & \textbf{85.2} & \textbf{88.6} & \textcolor{red}{\textbf{91.6}} & \textbf{86.8} & \textbf{74.3} & \textbf{64.3} & \textcolor{red}{\textbf{83.8}} & 97.9 & \textcolor{red}{\textbf{84.1}} \\
        \midrule
            \multicolumn{12}{l}{\textbf{Fine-tuned Models}} \\
        \midrule
            \multirow{2}{*}[-0.0ex]{\textbf{LLMs}} & \multirow{2}{*}[-0.0ex]{\textbf{Single}} & Llama-3.1-8B-Instruct & 75.4 & 79.6 & 89.3 & 77.6 & 69.6 & 61.7 & 79.7 & 97.7 & 78.8 \\
            & & Qwen2.5-7B-Instruct & 79.2 & 82.1 & 89.0 & 79.5 & 69.4 & 61.0 & 79.1 & 93.3 & 79.1 \\
        \midrule
            \textbf{MLLM} & \textbf{Single} & Qwen2-VL-7B-Instruct & 62.2 & 78.2 & 87.7 & 68.4 & 67.0 & 59.3 & 75.7 & 95.0 & 74.2 \\
        \bottomrule
        \end{tabular*}
    }
\end{table}

\section{Case Study}

\begin{figure}[H]
    \centering
    \includegraphics[width=1.0\linewidth]{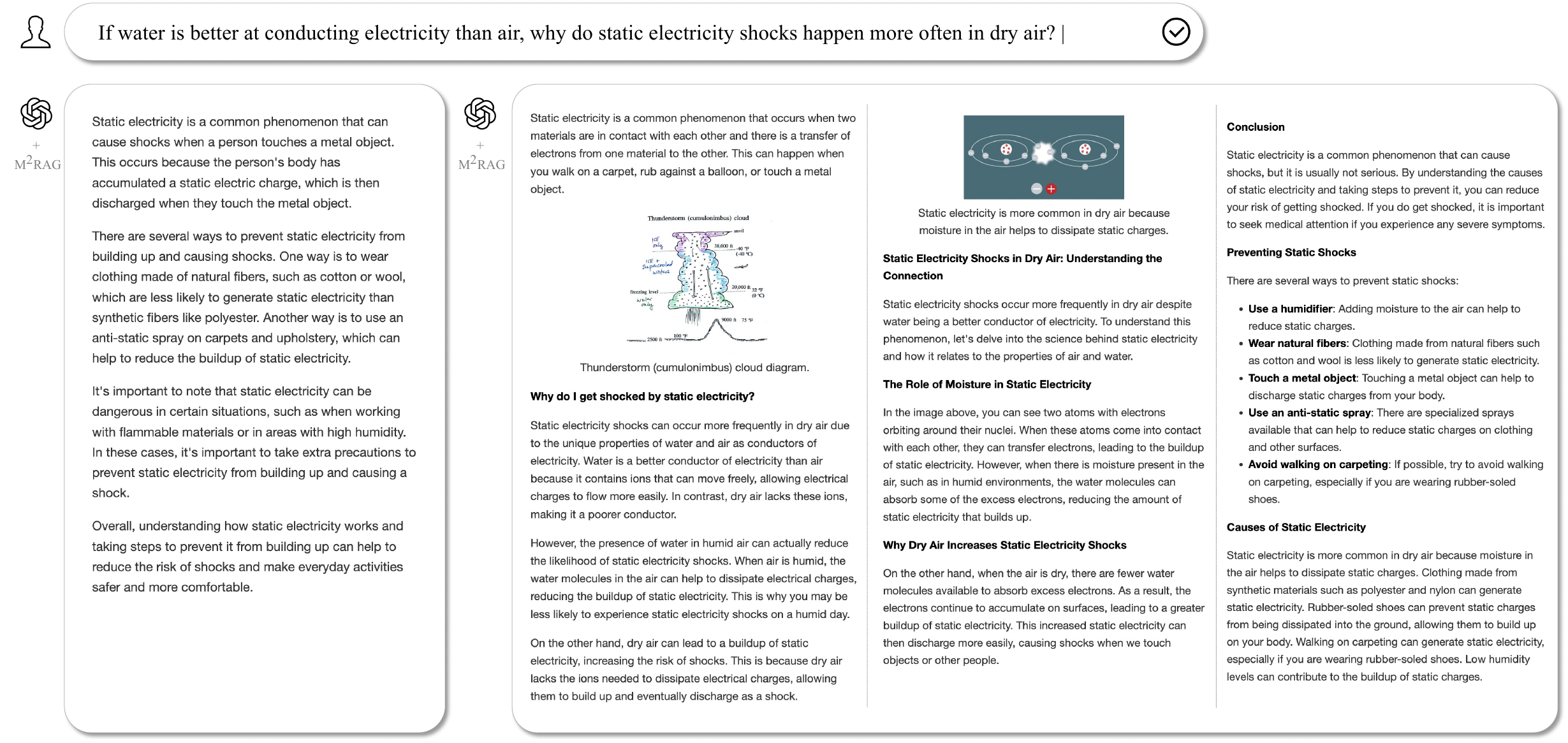}
    \caption{A case of comparison between single- and multi-stage approaches with Llama-3.2-Vision-90B-Instruct.}
    \label{fig:case_study_single_vs_multi_stage}
\end{figure}

\begin{figure}[H]
    \centering
    \includegraphics[width=1.0\linewidth]{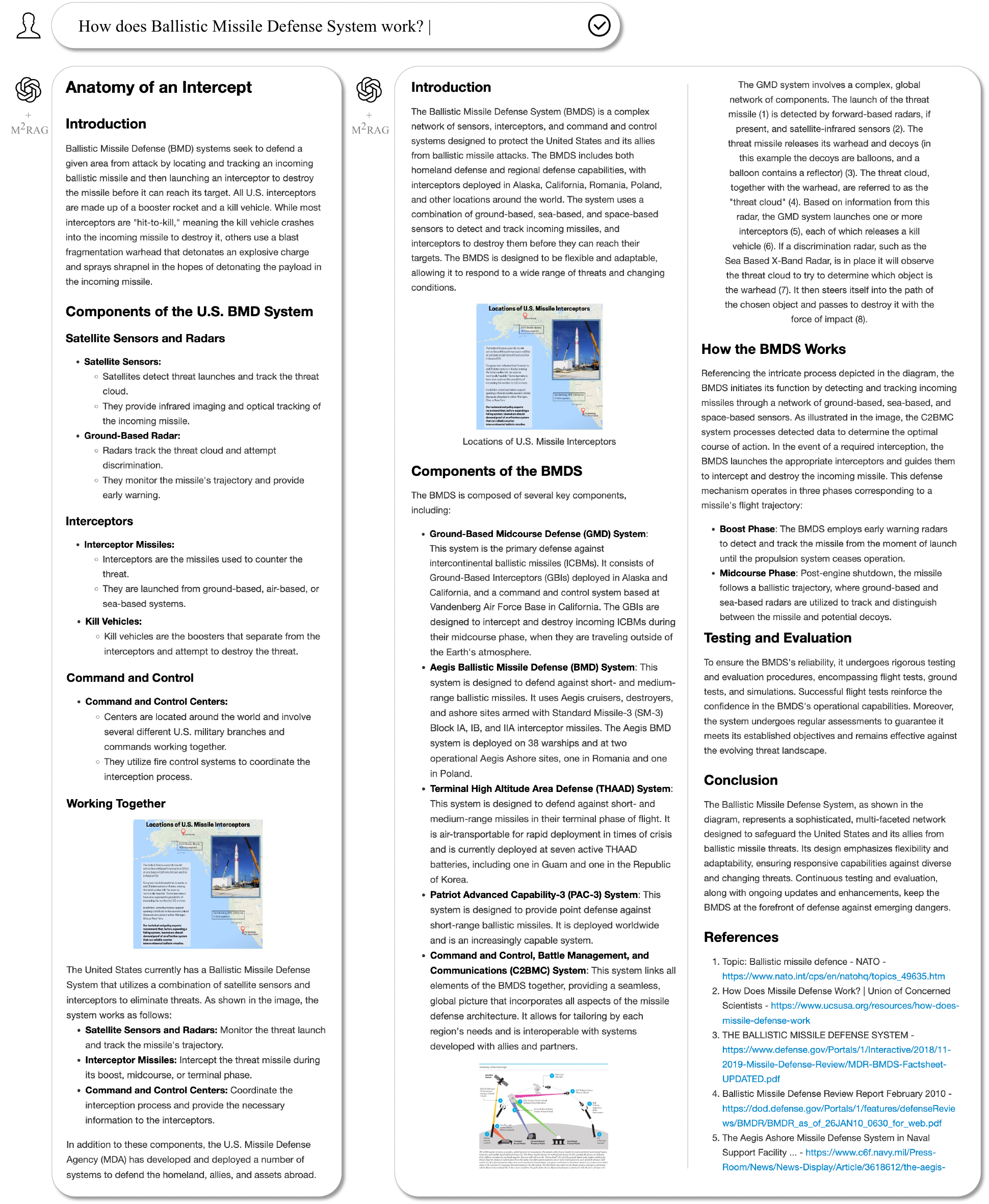}
    \caption{A case of comparison between small and large models (Qwen2-VL with 7B and 72B parameters).}
    \label{fig:case_study_small_vs_large_models}
\end{figure}

\begin{figure}[H]
    \centering
    \includegraphics[width=1.0\linewidth]{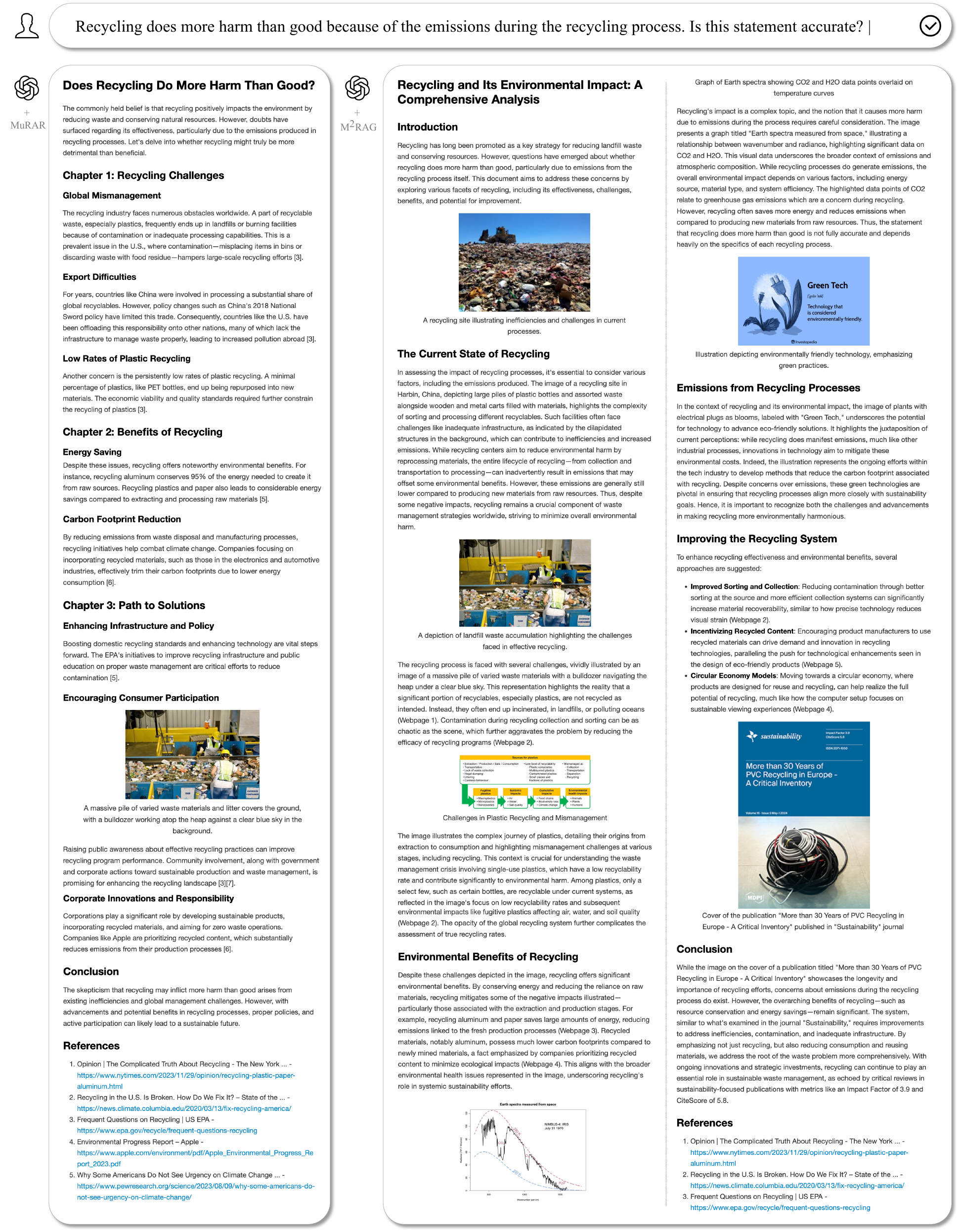}
    \caption{A case of comparison between separated and jointly modeling approaches with GPT-4o.}
    \label{fig:case_study_murar_vs_m2rag}
\end{figure}

In this section, we provide some cases to demonstrate more details in the M$^2$RAG task:

\begin{itemize}
    \item Single-stage versus Multi-stage, as illustrated in Figure~\ref{fig:case_study_single_vs_multi_stage};
    \item Large Models versus Small Models, as illustrated in Figure~\ref{fig:case_study_small_vs_large_models};
    \item Separate Modeling versus Joint Modeling, as illustrated in Figure~\ref{fig:case_study_murar_vs_m2rag}.
\end{itemize}

\end{document}